\documentclass[letterpaper]{article} 
\usepackage[submission]{aaai23}  
\usepackage{times}  
\usepackage{helvet}  
\usepackage{courier}  
\usepackage[hyphens]{url}  
\usepackage{graphicx} 
\usepackage{natbib}  
\usepackage{caption} 
\usepackage{algorithm}
\usepackage{algorithmic}

\usepackage{mathrsfs}
\usepackage{amssymb}
\usepackage{bm}
\usepackage{amsmath}
\usepackage{dsfont}
\usepackage{booktabs}
\usepackage{epstopdf}
\usepackage{multirow}
\usepackage{comment}
\usepackage[table,xcdraw]{xcolor}
\usepackage{cite}
\usepackage{endnotes}
\usepackage{makecell}

\usepackage[normalem]{ulem}
\useunder{\uline}{\ul}{}

\usepackage{newfloat}
\usepackage{listings}
\DeclareCaptionStyle{ruled}{labelfont=normalfont,labelsep=colon,strut=off} 
\floatstyle{ruled}
\newfloat{listing}{tb}{lst}{}
\floatname{listing}{Listing}

\title{Cluster-guided Contrastive Graph Clustering Network}
\author{Xihong Yang,\textsuperscript{\rm 1}\footnote{Equal contribution} Yue Liu,\textsuperscript{\rm 1}\footnotemark[1], Sihang Zhou,\textsuperscript{\rm 2} Siwei Wang,\textsuperscript{\rm 1} Wenxuan Tu,\textsuperscript{\rm 1} Qun Zheng,\textbf{\textsuperscript{\rm 3}}\\ Xinwang Liu,\textsuperscript{\rm 1}\footnote{Corresponding author} Liming Fang,\textsuperscript{\rm 4} En Zhu\textsuperscript{\rm 1}$^{\dagger}$}
\affiliations{
    \textsuperscript{\rm 1}College of Computer, National University of Defense Technology, Changsha, China\\
    \textsuperscript{\rm 2}College of Intelligence Science and Technology, National University of Defense Technology, Changsha, China\\
    \textsuperscript{\rm 3}University of Science and Technology of China,  \textsuperscript{\rm 4} Nanjing University of Aeronautics and Astronautics
    
    \{yangxihong, yueliu, wangsiwei13, twx, xinwangliu, enzhu\}@nudt.edu.cn, sihangjoe@gmail.com, zhengqun@mail.ustc.edu.cn, fangliming@nuaa.edu.cn
}

\usepackage{bibentry}

\begin{document}

\maketitle

\begin{abstract}

Benefiting from the intrinsic supervision information exploitation capability, contrastive learning has achieved promising performance in the field of deep graph clustering recently. However, we observe that two drawbacks of the positive and negative sample construction mechanisms limit the performance of existing algorithms from further improvement. 1) The quality of positive samples heavily depends on the carefully designed data augmentations, while inappropriate data augmentations would easily lead to the semantic drift and indiscriminative positive samples. 2) The constructed negative samples are not reliable for ignoring important clustering information. To solve these problems, we propose a Cluster-guided Contrastive deep Graph Clustering network (CCGC) by mining the intrinsic supervision information in the high-confidence clustering results. Specifically, instead of conducting complex node or edge perturbation, we construct two views of the graph by designing special Siamese encoders whose weights are not shared between the sibling sub-networks. Then, guided by the high-confidence clustering information, we carefully select and construct the positive samples from the same high-confidence cluster in two views. Moreover, to construct semantic meaningful negative sample pairs, we regard the centers of different high-confidence clusters as negative samples, thus improving the discriminative capability and reliability of the constructed sample pairs. Lastly, we design an objective function to pull close the samples from the same cluster while pushing away those from other clusters by maximizing and minimizing the cross-view cosine similarity between positive and negative samples. Extensive experimental results on six datasets demonstrate the effectiveness of CCGC compared with the existing state-of-the-art algorithms. The code of CCGC is available at https://github.com/xihongyang1999/CCGC on Github.

\end{abstract}

\begin{figure}[!t]
\centering
\small
\begin{minipage}{0.49\linewidth}
\centerline{\includegraphics[width=1\textwidth]{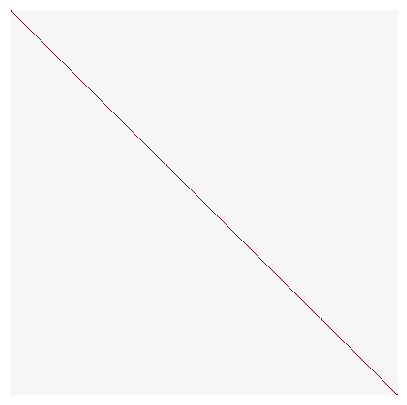}}
\vspace{3pt}
\textbf{\centerline{(a) GCA}}
\centerline{\includegraphics[width=1\textwidth]{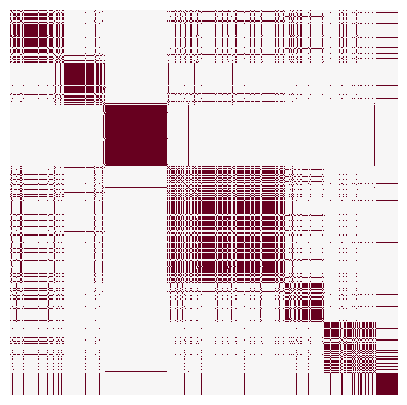}}
\vspace{3pt}
\textbf{\centerline{(c) Ours}}
\end{minipage}
\begin{minipage}{0.49\linewidth}
\centerline{\includegraphics[width=1\textwidth]{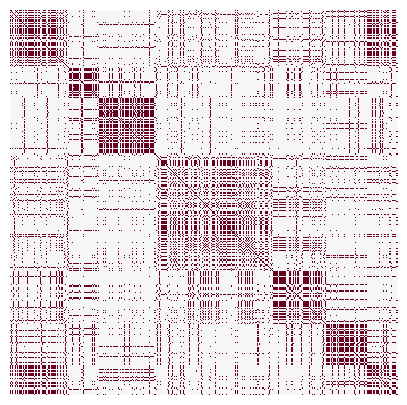}}
\vspace{3pt}
\textbf{\centerline{(b) SCAGC}}
\centerline{\includegraphics[width=\textwidth]{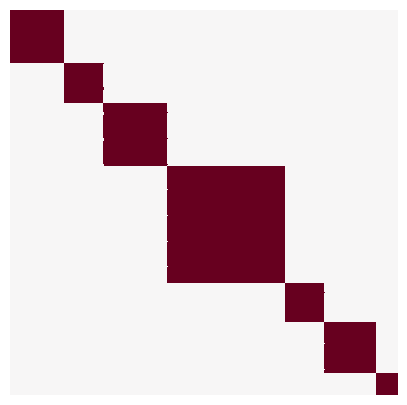}}
\vspace{3pt}
\textbf{\centerline{(d) Ground truth}}
\end{minipage}
\caption{Visualization of the positive sample pairs selected by (a) GCA \cite{GCA}, (b) SCAGC, \cite{SCAGC} and (c) the proposed method. The red dots denote the generated sample pairs. Specifically, if a point $(i,j)$ is selected as positive, the $i$-th sample from the first view and the $j$-th sample from the second view are integrated as a positive sample pair. (d) is the ground-truth cluster indicator. The sample order is rearranged to make samples from the same cluster beside each other. From the figures, we can find that our proposed positive sample extraction mechanism is more discriminative than the existing algorithms. As a consequence, the learned network is also more informative.}
\label{motivation}
\end{figure}

\section{Introduction}

Thanks to the strong representation learning capacity of the graph data, Graph Neural Networks (GNNs) have been successfully applied to various applications, such as node classification \cite{GCN,GAT,duan_nips,wangaug1, wangaug2,SUBLIME,xihong_MGCN}, graph classification \cite{yiqi_2,yiqi_3}, time series analysis\cite{liumeng_1,liumeng_2,xiefeng_1}, knowledge graph\cite{liangke,liangke_2}, and so on. Among all the directions in graph learning, deep graph clustering is a fundamental yet challenging unsupervised task, which has become a hot research spot recently \cite{DAEGC,DFCN,DCRN,yue_survey}.

Contrastive learning, which could capture the supervision information implicitly without human annotations, has become a prominent technique in deep graph clustering. Although promising performance has been achieved, we observe two issues in the contrastive sample-pair construction process. 1) The quality of positive samples heavily depends on the carefully selected graph data augmentations. However, inappropriate graph data augmentations, like random attribute permutation and random edge drop-out, would easily lead to semantic drift \cite{AFGRL}, and indiscriminative positive samples. 2) The constructed negative samples are not reliable enough since the existing algorithms neglect to exploit the important clustering information. Concretely, the existing methods randomly select negative samples, which loosely assign negative labels to samples from the same category. To improve the quality of negative samples, GDCL \cite{GDCL} and SCAGC \cite{SCAGC} randomly select samples from the different clusters. Although verified to be effective, the current clustering-result-based methods heavily rely on the carefully designed graph data augmentation and the well pre-trained model, thus limiting the clustering performance.

To solve these issues, we propose a novel Cluster-guided Contrastive deep Graph Clustering method, i.e., CCGC. Concretely, to construct two node views with different semantics, we take advantage of the Siamese encoders and make the parameters un-shared between two sub-networks. In this way, complex structure- and attribute-level data augmentations are avoided while the semantic drift problem has also been solved.

After that, we carefully select and construct the positive samples from the same cluster in two views according to high-confidence clustering pseudo labels. In this manner, we improve the discriminative capacity of the positive samples. As shown in Fig. \ref{motivation}, we visualize the positive sample pairs constructed by (a) GCA \cite{GCA}, (b) SCAGC \cite{SCAGC}, (c) our methods. It is clearly observed that our constructed positive samples could better reveal the ground truth compared to other methods. Meanwhile, we regard the centers of different high-confidence clusters as the negative sample pairs, which are more reliable and semantic meaningful. Moreover, we design an objective function to pull close the samples from the same cluster and push away those from different clusters by maximizing and minimizing the cross-view cosine similarity between positive and negative samples. The key contributions of this paper are listed as follows:

\begin{itemize}
\item We propose a cluster-guided contrastive deep graph clustering network termed CCGC to improve the quality of positive and negative samples by mining the high-confidence clustering information.

\item Instead of using carefully designed complex graph data augmentation, we conduct two views by designing special un-shared parameters Siamese encoders, thus avoiding semantic drift caused by inappropriate graph data augmentations.


\item Extensive experimental results on six benchmark datasets demonstrate the effectiveness of the proposed method against the existing state-of-the-art deep graph clustering competitors.

\end{itemize}

\begin{figure*}
\centering
\includegraphics[scale=0.4]{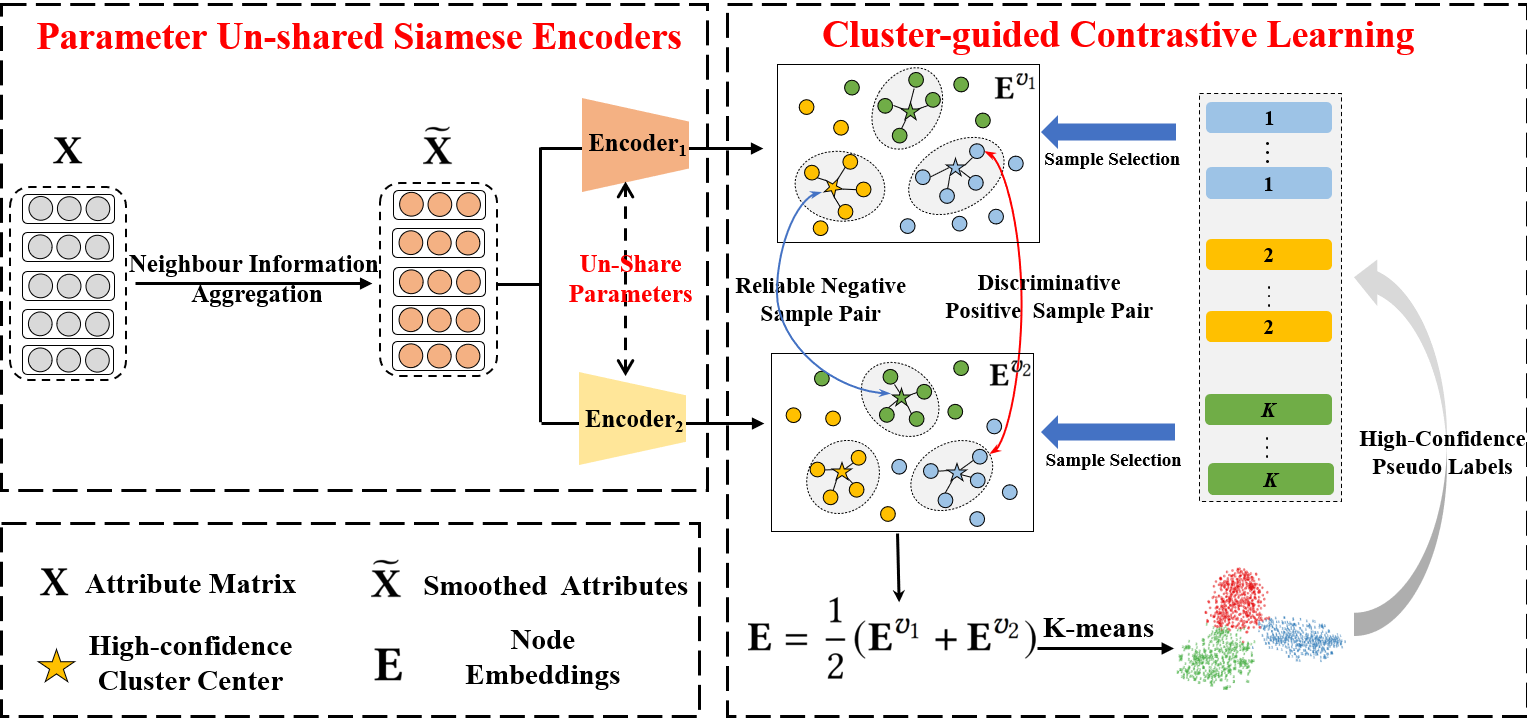}
\caption{Illustration of the Cluster-guided Contrastive Graph Clustering (CCGC) algorithm. In our proposed algorithm, we firstly encode the two-view node embeddings with the proposed parameter un-shared Siamese encoders. Then, we perform K-means on the fused node embeddings and obtain the clustering results. Subsequently, based on the high-confidence clustering results, we improve the quality of positive and negative samples by the discriminative positive sample construction strategy and reliable negative sample construction strategy in section \ref{strategy}. Lastly, we design an
objective function to pull close the samples from the same cluster while pushing away different high-confidence cluster centers, thus enhancing the discriminative capability of the network.}
\label{OVERRALL_FIGURE}  
\end{figure*}

\section{Related Work}
\subsection{Deep Graph Clustering}
Clustering is a fundamental yet challenging task, which aims to learn node semantic representations and divide nodes into different clusters \cite{suyuan,pei_1,pei_2,wanxinhang,xaiwei_2,Xiawei_1}. Deep learning methods also attract attention \cite{sihang_tip}. Among those methods, deep graph clustering has been a hot research spot in recent years.
According to the learning mechanism, the existing methods can be roughly grouped into three classes including generative methods, adversarial methods, and contrastive methods. Our survey paper \cite{yue_survey} summarizes the detailed information about the fast-growing deep graph clustering. CCGC is categorized into the last one, i.e., contrastive methods. This section reviews the generative methods and adversarial methods, and the contrastive methods will be detailed in the next section.

Inspired by the success of graph-auto-encoder (GAE) \cite{GAE}, the pioneer MGAE \cite{MGAE} first encodes the nodes with the graph-encoder \cite{GAE} and then performs clustering on the latent features. After that, DAEGC \cite{DAEGC} adopts the attention mechanisms \cite{attention,GAT} in early works to improve the clustering performance. Furthermore, ARGA \cite{ARGA} and AGAE \cite{AGAE} improve the discriminative capability of samples by adversarial mechanisms \cite{graph_GAN}. In addition, SDCN \cite{SDCN} alleviates the over-smoothing problem by integrating GAE and auto-encoder into the unified framework. More recently, R-GAE \cite{SCGC} enhances the existing GAE-based methods by alleviating the feature randomness and feature drift issues.

Although verified to be effective, since most of these methods adopt a distribution alignment loss function \cite{DEC} to force the learned node embeddings to have the minimum distortion against the pre-learned cluster centers, their clustering performance is highly dependent on good initial cluster centers, thus leading to manual trial-and-error pre-training \cite{DAEGC,ARGA,SDCN,DFCN}. As a consequence, the performance consistency, as well as the implementing convenience, is largely decreased. Different from them, several contrastive methods \cite{MVGRL,AGE,MCGC} replace the clustering guided loss function by the contrastive loss, thus getting rid of trial-and-error pre-training.

\subsection{Contrastive Deep Graph Clustering}
Contrastive learning has achieved great success in the fields of computer vision \cite{xihong} and graph learning \cite{simgrace,wangaug3,sail,huoshan} in recent years. Inspired by their success, contrastive deep graph clustering methods \cite{DCRN,IDCRN,GDCL} are increasingly proposed.

The fashions of the data augmentations and the positive-negative sample pair construction are two crucial factors to determine the performance of the contrastive deep graph clustering methods. In this section, we review the existing contrastive methods from these two perspectives.

\noindent{\textbf{Data augmentation.}} The technique of data augmentation plays an important role in contrastive deep graph clustering. Specifically, the existing methods construct different views of the graph by applying distinct augmentations to the graph. For example, the graph diffusion matrix would be regarded as one of the augmented graphs in MVGRL \cite{MVGRL}, GDCL \cite{GDCL}, and DCRN \cite{DCRN}. Differently, SCAGC \cite{SCAGC} randomly adds or drops edges to perturb the structure of graphs. From the feature perspective, DCRN and SCAGC conduct augmentations on node attributes by attribute corruption. Although verified to be effective, the promising performance of these methods highly depends on the carefully selected data augmentations. Some works \cite{AFGRL,MoCL} point out that the inappropriate data augmentations would easily lead to semantic drift and the similar conclusion could be found in section \ref{augmentation}. To overcome the issue, we propose a novel augmentation fashion to construct different graph views by setting the parameters of Siamese encoders to be un-shared, thus avoiding the semantic drift caused by inappropriate augmentations.

\noindent{\textbf{Positive and negative sample pair construction.}} Another crucial component in contrastive methods is the fashion of the positive and negative sample pair construction. Specifically, contrastive methods pull together positive samples while pushing away negative ones, thus the quality of positive and negative samples determines the performance of contrastive methods. Concretely, MVGRL \cite{MVGRL} regards different augmented views of the same node and generates the negative samples by randomly shuffling the feature. Besides, DCRN \cite{DCRN} pulls together the same node in different views while pushing away different nodes under the feature decorrelation constrain. Moreover, SCAGC \cite{SCAGC} and GDCL \cite{GDCL} improve the quality of negative samples by randomly selecting samples from the different clusters. Although verified to be effective, they still rely on a well pre-trained model to select high-quality positive and negative samples. To solve this problem, we propose a high-confidence clustering information guided fashion of positive and negative sample construction, thus enhancing the discriminative capability and reliability of the sample pairs.

\section{Method}
In this section, we propose a novel Cluster-guided Contrastive deep Graph Clustering algorithm (CCGC). The overall framework of CCGC is shown in Fig. \ref{OVERRALL_FIGURE}. In the following sections, we will introduce the proposed CCGC in specific.


\subsection{Notations and Problem Definition}
In an undirected graph $\mathcal{G}=\left \{\textbf{X}, \textbf{A} \right \}$, let $\mathcal{V}=\{v_1, v_2, \dots, v_N\}$ be a set of $N$ nodes with $K$ classes and $\mathcal{E}$ be a set of edges. $\textbf{X} \in \mathds{R}^{N\times D}$ and $\textbf{A} \in \mathds{R}^{N\times N}$ denote the attribute matrix and the original adjacency matrix, respectively. The degree matrix is formulated as $\textbf{D}=diag(d_1, d_2, \dots ,d_N)\in \mathds{R}^{N\times N}$ and $d_i=\sum_{(v_i,v_j)\in \mathcal{E}}a_{ij}$. The graph Laplacian matrix is defined as $\textbf{L}=\textbf{D}-\textbf{A}$. With the renormalization trick $\widehat{\textbf{A}} = \textbf{A} + \textbf{I}$, the symmetric normalized graph Laplacian matrix is denoted as $\widetilde{\textbf{L}} = \textbf{I} - \widehat{\textbf{D}}^{-\frac{1}{2}}\widehat{\textbf{A}}\widehat{\textbf{D}}^{-\frac{1}{2}}$. 

\subsection{Parameter Un-shared Siamese Encoders}
In this section, following SCGC \cite{SCGC}, we embed the nodes into the latent space and construct two different sample views by designing a parameter un-shared Siamese encoders.

Before encoding, we adopt a widely-used Laplacian filter \cite{AGE} to conduct neighbour information aggregation as follows:
\begin{equation} 
\widetilde{\textbf{X}} = (\textbf{I}-\widetilde{\textbf{L}})^t\textbf{X},
\label{filter}
\end{equation}
where $\widetilde{\textbf{L}}$ is the symmetric normalized graph Laplacian matrix. $t$ denotes the layer number of the Laplacian filter. $\widetilde{\textbf{X}}$ is the smoothed attribute matrix. Then we encode $\widetilde{\textbf{X}}$ with MLP encoders as follows:
\begin{equation} 
\textbf{E}^{v_1} = \text{Encoder}_1(\widetilde{\textbf{X}}), \ 
\textbf{E}^{v_2} = \text{Encoder}_2(\widetilde{\textbf{X}}), 
\label{ENCODER}
\end{equation}
where $\textbf{E}^{v_1}$ and $\textbf{E}^{v_2}$ denotes the first and second view of the node embeddings, respectively. For the encoders, we design them to have the same architecture but un-shared learnable parameters. Subsequently, we normalize $\textbf{E}^{v_1}$ and $\textbf{E}^{v_2}$ with ${\ell ^2}$-norm:
\begin{equation} 
\textbf{E}^{v_1} = \frac{\textbf{E}^{v_1}}{||\textbf{E}^{v_1}||_2},\
\textbf{E}^{v_2} = \frac{\textbf{E}^{v_2}}{||\textbf{E}^{v_2}||_2}.
\label{normal}
\end{equation}

By this setting, we construct two node views with different semantic, thus avoiding semantic drift caused by inappropriate data augmentations on the graphs. Experimental evidence could be found in section \ref{augmentation}.

\subsection{Cluster-guided Contrastive Learning}
\label{strategy}
In this section, we propose the Cluster-guided Contrastive Learning (CCL) to improve the discriminative capability and reliability of samples by mining the high-confidence clustering information.

To be specific, we firstly fuse the two views of the node embeddings as follows:
\begin{equation}
\textbf{E} = \frac{1}{2}(\textbf{E}^{v_1}+\textbf{E}^{v_2}).
\label{FUSION}
\end{equation}
Then we perform K-means on $\textbf{E}$ and obtain the clustering results. In order to generate more reliable clustering information \cite{IDCRN}, we define the confidence score $\textbf{CONF}_i$ of $i$-th sample as formulated:
\begin{equation}
\textbf{CONF}_i = e^{-||\textbf{E}_i-\textbf{C}_p||^2},
\label{high_confidence}
\end{equation}
where $\textbf{E}_i$ denotes the $i$-th node embedding. Besides, $\textbf{C}_p (p = 1,2,...,K)$ denotes the center of the cluster, which contains the $i$-th sample. Subsequently, based on $\textbf{CONF}$, we denote the high-confidence sample indexes $h$ as follows:

\begin{equation}
h = \{h_1, h_2, ..., h_i, ...\},
\label{index}
\end{equation}
where the element $h_i$ indicates that $h_i$-th sample belongs to top $\tau$ high-confidence sample set.

Based on these high-confidence samples and their clustering pseudo labels, we propose two sample construction strategies including Discriminative Positive sample construction Strategy (DPS) and Reliable Negative sample construction Strategy (RNS).

\noindent{\textbf{Discriminative Positive Sample Construction Strategy.}}
In this part, we design DPS to enhance the discriminative capability of positive samples. The proposed DPS contains three steps. Firstly, we select the high-confidence samples of two views with high-confidence indexes $h$ as follows:
\begin{equation}
\textbf{H}^{v_1} = \textbf{E}^{v_1}_{[h,:]}, \
\textbf{H}^{v_2} = \textbf{E}^{v_2}_{[h,:]}.
\label{two view}
\end{equation}
Then, according to the corresponding pseudo labels, we group $\textbf{H}^{v_1}$ and $\textbf{H}^{v_2}$ into $K$ disjoint clusters, i.e., $\textbf{B}^{v_1}_p ( p=1,2,...,K)$, and $\textbf{B}^{v_2}_q (q=1,2,...,K)$. Subsequently, the positive samples will be selected and constructed from the same high-confidence clusters in Eq. \eqref{pos_loss}. In this setting, the high-confidence clustering pseudo labels could be utilized as the supervisory information to improve the discriminative capability of the positive samples.

\noindent{\textbf{Reliable Negative Sample Construction Strategy.}} For the negative sample construction, the existing works \cite{GCA,DCRN} directly regard all other non-positive samples as negative samples, easily bringing false-negative samples. To alleviate this issue, we propose RNS, which contains two steps. Concretely, we first calculate the centers of high-confidence samples in two views:

\begin{equation}
\begin{aligned}
\textbf{CEN}^{v_1}_p &= avg(\textbf{B}^{v_1}_p), p=1,2,...,K,\\
\textbf{CEN}^{v_2}_q &= avg(\textbf{B}^{v_2}_q), q=1,2,...,K,
\end{aligned}
\label{neg_center}
\end{equation}
where $avg$ is the average function. Then we regard the different high-confidence centers as the negative samples in Eq. \eqref{neg_loss}. In this manner, RNS would enhance the reliability of negative samples, thus reducing the possibility of false-negative samples.

In summary, the proposed CCL would guide our network to mine the supervisory information in the high-confidence clustering pseudo labels, thus improving the discriminative capability and reliability of samples. 

\begin{table*}[]
\centering
\setlength{\tabcolsep}{4pt}
\scalebox{0.63}{
\begin{tabular}{
>{\columncolor[HTML]{FFFFFF}}c |
>{\columncolor[HTML]{FFFFFF}}c |
>{\columncolor[HTML]{FFFFFF}}c 
>{\columncolor[HTML]{FFFFFF}}c 
>{\columncolor[HTML]{FFFFFF}}c 
>{\columncolor[HTML]{FFFFFF}}c 
>{\columncolor[HTML]{FFFFFF}}c 
>{\columncolor[HTML]{FFFFFF}}c 
>{\columncolor[HTML]{FFFFFF}}c 
>{\columncolor[HTML]{FFFFFF}}c 
>{\columncolor[HTML]{FFFFFF}}c 
>{\columncolor[HTML]{FFFFFF}}c 
>{\columncolor[HTML]{FFFFFF}}c 
>{\columncolor[HTML]{FFFFFF}}c |
>{\columncolor[HTML]{FFFFFF}}c }
\hline
\multicolumn{1}{l|}{\cellcolor[HTML]{FFFFFF}{\color[HTML]{000000} }}                                   & \multicolumn{1}{l|}{\cellcolor[HTML]{FFFFFF}{\color[HTML]{000000} }}                                  & {\color[HTML]{000000} \textbf{MGAE}}    & {\color[HTML]{000000} \textbf{DAEGC}}    & {\color[HTML]{000000} \textbf{ARGA}}     & {\color[HTML]{000000} \textbf{SDCN}}   & {\color[HTML]{000000} \textbf{DFCN}}    & {\color[HTML]{000000} \textbf{AGE}}       & {\color[HTML]{000000} \textbf{MVGRL}}   & {\color[HTML]{000000} \textbf{AutoSSL}} & {\color[HTML]{000000} \textbf{AGC-DRR}}  & {\color[HTML]{000000} \textbf{AFGRL}}   & {\color[HTML]{000000} \textbf{GDCL}}     & {\color[HTML]{000000} \textbf{ProGCL}}  & {\color[HTML]{000000} \textbf{CCGC}} \\ \cline{3-15} 
\multicolumn{1}{l|}{\multirow{-2}{*}{\cellcolor[HTML]{FFFFFF}{\color[HTML]{000000} \textbf{Dataset}}}} & \multicolumn{1}{l|}{\multirow{-2}{*}{\cellcolor[HTML]{FFFFFF}{\color[HTML]{000000} \textbf{Metrix}}}} & {\color[HTML]{000000} \textbf{CIKM 17}} & {\color[HTML]{000000} \textbf{IJCAI 19}} & {\color[HTML]{000000} \textbf{IJCAI 19}} & {\color[HTML]{000000} \textbf{WWW 20}} & {\color[HTML]{000000} \textbf{AAAI 21}} & {\color[HTML]{000000} \textbf{SIGKDD 20}} & {\color[HTML]{000000} \textbf{ICML 20}} & {\color[HTML]{000000} \textbf{ICLR 22}} & {\color[HTML]{000000} \textbf{IJCAI 22}} & {\color[HTML]{000000} \textbf{AAAI 22}} & {\color[HTML]{000000} \textbf{IJCAI 21}} & {\color[HTML]{000000} \textbf{ICML 22}} & {\color[HTML]{000000} \textbf{Ours}} \\ \hline
\cellcolor[HTML]{FFFFFF}{\color[HTML]{000000} }                                                        & {\color[HTML]{000000} ACC}                                                                            & {\color[HTML]{000000} 43.38±2.11}       & {\color[HTML]{000000} 70.43±0.36}        & {\color[HTML]{000000} 71.04±0.25}        & {\color[HTML]{000000} 35.60±2.83}      & {\color[HTML]{000000} 36.33±0.49}       & {\color[HTML]{0000FF} 73.50±1.83}         & {\color[HTML]{000000} 70.47±3.70}       & {\color[HTML]{000000} 63.81±0.57}       & {\color[HTML]{000000} 40.62±0.55}        & {\color[HTML]{000000} 26.25±1.24}       & {\color[HTML]{000000} 70.83±0.47}        & {\color[HTML]{000000} 57.13+1.23}       & {\color[HTML]{FF0000} 73.88±1.20}    \\
\cellcolor[HTML]{FFFFFF}{\color[HTML]{000000} }                                                        & {\color[HTML]{000000} NMI}                                                                            & {\color[HTML]{000000} 28.78±2.97}       & {\color[HTML]{000000} 52.89±0.69}        & {\color[HTML]{000000} 51.06±0.52}        & {\color[HTML]{000000} 14.28±1.91}      & {\color[HTML]{000000} 19.36±0.87}       & {\color[HTML]{FF0000} 57.58±1.42}         & {\color[HTML]{000000} 55.57±1.54}       & {\color[HTML]{000000} 47.62±0.45}       & {\color[HTML]{000000} 18.74±0.73}        & {\color[HTML]{000000} 12.36±1.54}       & {\color[HTML]{000000} 56.30±0.36}        & {\color[HTML]{000000} 41.02+1.34}       & {\color[HTML]{0000FF} 56.45±1.04}    \\
\cellcolor[HTML]{FFFFFF}{\color[HTML]{000000} }                                                        & {\color[HTML]{000000} ARI}                                                                            & {\color[HTML]{000000} 16.43±1.65}       & {\color[HTML]{000000} 49.63±0.43}        & {\color[HTML]{000000} 47.71±0.33}        & {\color[HTML]{000000} 07.78±3.24}      & {\color[HTML]{000000} 04.67±2.10}       & {\color[HTML]{0000FF} 50.10±2.14}         & {\color[HTML]{000000} 48.70±3.94}       & {\color[HTML]{000000} 38.92±0.77}       & {\color[HTML]{000000} 14.80±1.64}        & {\color[HTML]{000000} 14.32±1.87}       & {\color[HTML]{000000} 48.05±0.72}        & {\color[HTML]{000000} 30.71+2.70}       & {\color[HTML]{FF0000} 52.51±1.89}    \\
\multirow{-4}{*}{\cellcolor[HTML]{FFFFFF}{\color[HTML]{000000} \textbf{CORA}}}                         & {\color[HTML]{000000} F1}                                                                             & {\color[HTML]{000000} 33.48±3.05}       & {\color[HTML]{000000} 68.27±0.57}        & {\color[HTML]{000000} 69.27±0.39}        & {\color[HTML]{000000} 24.37±1.04}      & {\color[HTML]{000000} 26.16±0.50}       & {\color[HTML]{0000FF} 69.28±1.59}         & {\color[HTML]{000000} 67.15±1.86}       & {\color[HTML]{000000} 56.42±0.21}       & {\color[HTML]{000000} 31.23±0.57}        & {\color[HTML]{000000} 30.20±1.15}       & {\color[HTML]{000000} 52.88±0.97}        & {\color[HTML]{000000} 45.68+1.29}       & {\color[HTML]{FF0000} 70.98±2.79}    \\ \hline
\cellcolor[HTML]{FFFFFF}{\color[HTML]{000000} }                                                        & {\color[HTML]{000000} ACC}                                                                            & {\color[HTML]{000000} 61.35±0.80}       & {\color[HTML]{000000} 64.54±1.39}        & {\color[HTML]{000000} 61.07±0.49}        & {\color[HTML]{000000} 65.96±0.31}      & {\color[HTML]{000000} 69.50±0.20}       & {\color[HTML]{0000FF} 69.73±0.24}         & {\color[HTML]{000000} 62.83±1.59}       & {\color[HTML]{000000} 66.76±0.67}       & {\color[HTML]{000000} 68.32±1.83}        & {\color[HTML]{000000} 31.45±0.54}       & {\color[HTML]{000000} 66.39±0.65}        & {\color[HTML]{000000} 65.92+0.80}       & {\color[HTML]{FF0000} 69.84±0.94}    \\
\cellcolor[HTML]{FFFFFF}{\color[HTML]{000000} }                                                        & {\color[HTML]{000000} NMI}                                                                            & {\color[HTML]{000000} 34.63±0.65}       & {\color[HTML]{000000} 36.41±0.86}        & {\color[HTML]{000000} 34.40±0.71}        & {\color[HTML]{000000} 38.71±0.32}      & {\color[HTML]{000000} 43.90±0.20}       & {\color[HTML]{FF0000} 44.93±0.53}         & {\color[HTML]{000000} 40.69±0.93}       & {\color[HTML]{000000} 40.67±0.84}       & {\color[HTML]{000000} 43.28±1.41}        & {\color[HTML]{000000} 15.17±0.47}       & {\color[HTML]{000000} 39.52±0.38}        & {\color[HTML]{000000} 39.59+0.39}       & {\color[HTML]{0000FF} 44.33±0.79}    \\
\cellcolor[HTML]{FFFFFF}{\color[HTML]{000000} }                                                        & {\color[HTML]{000000} ARI}                                                                            & {\color[HTML]{000000} 33.55±1.18}       & {\color[HTML]{000000} 37.78±1.24}        & {\color[HTML]{000000} 34.32±0.70}        & {\color[HTML]{000000} 40.17±0.43}      & {\color[HTML]{0000FF} 45.50±0.30}       & {\color[HTML]{000000} 45.31±0.41}         & {\color[HTML]{000000} 34.18±1.73}       & {\color[HTML]{000000} 38.73±0.55}       & {\color[HTML]{000000} 45.34±2.33}        & {\color[HTML]{000000} 14.32±0.78}       & {\color[HTML]{000000} 41.07±0.96}        & {\color[HTML]{000000} 36.16+1.11}       & {\color[HTML]{FF0000} 45.68±1.80}    \\
\multirow{-4}{*}{\cellcolor[HTML]{FFFFFF}{\color[HTML]{000000} \textbf{CITESEER}}}                     & {\color[HTML]{000000} F1}                                                                             & {\color[HTML]{000000} 57.36±0.82}       & {\color[HTML]{000000} 62.20±1.32}        & {\color[HTML]{000000} 58.23±0.31}        & {\color[HTML]{000000} 63.62±0.24}      & {\color[HTML]{000000} 64.30±0.20}       & {\color[HTML]{0000FF} 64.45±0.27}         & {\color[HTML]{000000} 59.54±2.17}       & {\color[HTML]{000000} 58.22±0.68}       & {\color[HTML]{FF0000} 64.82±1.60}        & {\color[HTML]{000000} 30.20±0.71}       & {\color[HTML]{000000} 61.12±0.70}        & {\color[HTML]{000000} 57.89+1.98}       & {\color[HTML]{000000} 62.71±2.06}    \\ \hline
\cellcolor[HTML]{FFFFFF}{\color[HTML]{000000} }                                                        & {\color[HTML]{000000} ACC}                                                                            & {\color[HTML]{000000} 71.57±2.48}       & {\color[HTML]{000000} 75.96±0.23}        & {\color[HTML]{000000} 69.28±2.30}        & {\color[HTML]{000000} 53.44±0.81}      & {\color[HTML]{0000FF} 76.82±0.23}       & {\color[HTML]{000000} 75.98±0.68}         & {\color[HTML]{000000} 41.07±3.12}       & {\color[HTML]{000000} 54.55±0.97}       & {\color[HTML]{000000} 76.81±1.45}        & {\color[HTML]{000000} 75.51±0.77}       & {\color[HTML]{000000} 43.75±0.78}        & {\color[HTML]{000000} 51.53+0.38}       & {\color[HTML]{FF0000} 77.25±0.41}    \\
\cellcolor[HTML]{FFFFFF}{\color[HTML]{000000} }                                                        & {\color[HTML]{000000} NMI}                                                                            & {\color[HTML]{000000} 62.13±2.79}       & {\color[HTML]{000000} 65.25±0.45}        & {\color[HTML]{000000} 58.36±2.76}        & {\color[HTML]{000000} 44.85±0.83}      & {\color[HTML]{000000} 66.23±1.21}       & {\color[HTML]{000000} 65.38±0.61}         & {\color[HTML]{000000} 30.28±3.94}       & {\color[HTML]{000000} 48.56±0.71}       & {\color[HTML]{0000FF} 66.54±1.24}        & {\color[HTML]{000000} 64.05±0.15}       & {\color[HTML]{000000} 37.32±0.28}        & {\color[HTML]{000000} 39.56+0.39}       & {\color[HTML]{FF0000} 67.44±0.48}    \\
\cellcolor[HTML]{FFFFFF}{\color[HTML]{000000} }                                                        & {\color[HTML]{000000} ARI}                                                                            & {\color[HTML]{000000} 48.82±4.57}       & {\color[HTML]{000000} 58.12±0.24}        & {\color[HTML]{000000} 44.18±4.41}        & {\color[HTML]{000000} 31.21±1.23}      & {\color[HTML]{0000FF} 58.28±0.74}       & {\color[HTML]{000000} 55.89±1.34}         & {\color[HTML]{000000} 18.77±2.34}       & {\color[HTML]{000000} 26.87±0.34}       & {\color[HTML]{FF0000} 60.15±1.56}        & {\color[HTML]{000000} 54.45±0.48}       & {\color[HTML]{000000} 21.57±0.51}        & {\color[HTML]{000000} 34.18+0.89}       & {\color[HTML]{000000} 57.99±0.66}    \\
\multirow{-4}{*}{\cellcolor[HTML]{FFFFFF}{\color[HTML]{000000} \textbf{AMAP}}}                         & {\color[HTML]{000000} F1}                                                                             & {\color[HTML]{000000} 68.08±1.76}       & {\color[HTML]{000000} 69.87±0.54}        & {\color[HTML]{000000} 64.30±1.95}        & {\color[HTML]{000000} 50.66±1.49}      & {\color[HTML]{000000} 71.25±0.31}       & {\color[HTML]{FF0000} 71.74±0.93}         & {\color[HTML]{000000} 32.88±5.50}       & {\color[HTML]{000000} 54.47±0.83}       & {\color[HTML]{000000} 71.03±0.64}        & {\color[HTML]{000000} 69.99±0.34}       & {\color[HTML]{000000} 38.37±0.29}        & {\color[HTML]{000000} 31.97+0.44}       & {\color[HTML]{FF0000} 72.18±0.57}    \\ \hline
\cellcolor[HTML]{FFFFFF}{\color[HTML]{000000} }                                                        & {\color[HTML]{000000} ACC}                                                                            & {\color[HTML]{000000} 53.59±2.04}       & {\color[HTML]{000000} 52.67±0.00}        & {\color[HTML]{0000FF} 67.86±0.80}        & {\color[HTML]{000000} 53.05±4.63}      & {\color[HTML]{000000} 55.73±0.06}       & {\color[HTML]{000000} 56.68±0.76}         & {\color[HTML]{000000} 37.56±0.32}       & {\color[HTML]{000000} 42.43±0.47}       & {\color[HTML]{000000} 47.79±0.02}        & {\color[HTML]{000000} 50.92±0.44}       & {\color[HTML]{000000} 45.42±0.54}        & {\color[HTML]{000000} 55.73+0.79}       & {\color[HTML]{FF0000} 75.04±1.78}    \\
\cellcolor[HTML]{FFFFFF}{\color[HTML]{000000} }                                                        & {\color[HTML]{000000} NMI}                                                                            & {\color[HTML]{000000} 30.59±2.06}       & {\color[HTML]{000000} 21.43±0.35}        & {\color[HTML]{0000FF} 49.09±0.54}        & {\color[HTML]{000000} 25.74±5.71}      & {\color[HTML]{000000} 48.77±0.51}       & {\color[HTML]{000000} 36.04±1.54}         & {\color[HTML]{000000} 29.33±0.70}       & {\color[HTML]{000000} 17.84±0.98}       & {\color[HTML]{000000} 19.91±0.24}        & {\color[HTML]{000000} 27.55±0.62}       & {\color[HTML]{000000} 31.70±0.42}        & {\color[HTML]{000000} 28.69+0.92}       & {\color[HTML]{FF0000} 50.23±2.43}    \\
\cellcolor[HTML]{FFFFFF}{\color[HTML]{000000} }                                                        & {\color[HTML]{000000} ARI}                                                                            & {\color[HTML]{000000} 24.15±1.70}       & {\color[HTML]{000000} 18.18±0.29}        & {\color[HTML]{0000FF} 42.02±1.21}        & {\color[HTML]{000000} 21.04±4.97}      & {\color[HTML]{000000} 37.76±0.23}       & {\color[HTML]{000000} 26.59±1.83}         & {\color[HTML]{000000} 13.45±0.03}       & {\color[HTML]{000000} 13.11±0.81}       & {\color[HTML]{000000} 14.59±0.13}        & {\color[HTML]{000000} 21.89±0.74}       & {\color[HTML]{000000} 19.33±0.57}        & {\color[HTML]{000000} 21.84+1.34}       & {\color[HTML]{FF0000} 46.95±3.09}    \\
\multirow{-4}{*}{\cellcolor[HTML]{FFFFFF}{\color[HTML]{000000} \textbf{BAT}}}                          & {\color[HTML]{000000} F1}                                                                             & {\color[HTML]{000000} 50.83±3.23}       & {\color[HTML]{000000} 52.23±0.03}        & {\color[HTML]{0000FF} 67.02±1.15}        & {\color[HTML]{000000} 46.45±5.90}      & {\color[HTML]{000000} 50.90±0.12}       & {\color[HTML]{000000} 55.07±0.80}         & {\color[HTML]{000000} 29.64±0.49}       & {\color[HTML]{000000} 34.84±0.15}       & {\color[HTML]{000000} 42.33±0.51}        & {\color[HTML]{000000} 46.53±0.57}       & {\color[HTML]{000000} 39.94±0.57}        & {\color[HTML]{000000} 56.08+0.89}       & {\color[HTML]{FF0000} 74.90±1.80}    \\ \hline
\cellcolor[HTML]{FFFFFF}{\color[HTML]{000000} }                                                        & {\color[HTML]{000000} ACC}                                                                            & {\color[HTML]{000000} 44.61±2.10}       & {\color[HTML]{000000} 36.89±0.15}        & {\color[HTML]{0000FF} 52.13±0.00}        & {\color[HTML]{000000} 39.07±1.51}      & {\color[HTML]{000000} 49.37±0.19}       & {\color[HTML]{000000} 47.26±0.32}         & {\color[HTML]{000000} 32.88±0.71}       & {\color[HTML]{000000} 31.33±0.52}       & {\color[HTML]{000000} 37.37±0.11}        & {\color[HTML]{000000} 37.42±1.24}       & {\color[HTML]{000000} 33.46±0.18}        & {\color[HTML]{000000} 43.36+0.87}       & {\color[HTML]{FF0000} 57.19±0.66}    \\
\cellcolor[HTML]{FFFFFF}{\color[HTML]{000000} }                                                        & {\color[HTML]{000000} NMI}                                                                            & {\color[HTML]{000000} 15.60±2.30}       & {\color[HTML]{000000} 05.57±0.06}        & {\color[HTML]{000000} 22.48±1.21}        & {\color[HTML]{000000} 08.83±2.54}      & {\color[HTML]{000000} 32.90±0.41}       & {\color[HTML]{000000} 23.74±0.90}         & {\color[HTML]{000000} 11.72±1.08}       & {\color[HTML]{000000} 07.63±0.85}       & {\color[HTML]{000000} 07.00±0.85}        & {\color[HTML]{000000} 11.44±1.41}       & {\color[HTML]{000000} 13.22±0.33}        & {\color[HTML]{0000FF} 23.93+0.45}       & {\color[HTML]{FF0000} 33.85±0.87}    \\
\cellcolor[HTML]{FFFFFF}{\color[HTML]{000000} }                                                        & {\color[HTML]{000000} ARI}                                                                            & {\color[HTML]{000000} 13.40±1.26}       & {\color[HTML]{000000} 05.03±0.08}        & {\color[HTML]{000000} 17.29±0.50}        & {\color[HTML]{000000} 06.31±1.95}      & {\color[HTML]{0000FF} 23.25±0.18}       & {\color[HTML]{000000} 16.57±0.46}         & {\color[HTML]{000000} 04.68±1.30}       & {\color[HTML]{000000} 02.13±0.67}       & {\color[HTML]{000000} 04.88±0.91}        & {\color[HTML]{000000} 06.57±1.73}       & {\color[HTML]{000000} 04.31±0.29}        & {\color[HTML]{000000} 15.03+0.98}       & {\color[HTML]{FF0000} 27.71±0.41}    \\
\multirow{-4}{*}{\cellcolor[HTML]{FFFFFF}{\color[HTML]{000000} \textbf{EAT}}}                          & {\color[HTML]{000000} F1}                                                                             & {\color[HTML]{000000} 43.08±3.26}       & {\color[HTML]{000000} 34.72±0.16}        & {\color[HTML]{0000FF} 52.75±0.07}        & {\color[HTML]{000000} 33.42±3.10}      & {\color[HTML]{000000} 42.95±0.04}       & {\color[HTML]{000000} 45.54±0.40}         & {\color[HTML]{000000} 25.35±0.75}       & {\color[HTML]{000000} 21.82±0.98}       & {\color[HTML]{000000} 35.20±0.17}        & {\color[HTML]{000000} 30.53±1.47}       & {\color[HTML]{000000} 25.02±0.21}        & {\color[HTML]{000000} 42.54+0.45}       & {\color[HTML]{FF0000} 57.09±0.94}    \\ \hline
\cellcolor[HTML]{FFFFFF}{\color[HTML]{000000} }                                                        & {\color[HTML]{000000} ACC}                                                                            & {\color[HTML]{000000} 48.97±1.52}       & {\color[HTML]{000000} 52.29±0.49}        & {\color[HTML]{000000} 49.31±0.15}        & {\color[HTML]{000000} 52.25±1.91}      & {\color[HTML]{000000} 33.61±0.09}       & {\color[HTML]{0000FF} 52.37±0.42}         & {\color[HTML]{000000} 44.16±1.38}       & {\color[HTML]{000000} 42.52±0.64}       & {\color[HTML]{000000} 42.64±0.31}        & {\color[HTML]{000000} 41.50±0..25}      & {\color[HTML]{000000} 48.70±0.06}        & {\color[HTML]{000000} 45.38+0.58}       & {\color[HTML]{FF0000} 56.34±1.11}    \\
\cellcolor[HTML]{FFFFFF}{\color[HTML]{000000} }                                                        & {\color[HTML]{000000} NMI}                                                                            & {\color[HTML]{000000} 20.69±0.98}       & {\color[HTML]{000000} 21.33±0.44}        & {\color[HTML]{0000FF} 25.44±0.31}        & {\color[HTML]{000000} 21.61±1.26}      & {\color[HTML]{000000} 26.49±0.41}       & {\color[HTML]{000000} 23.64±0.66}         & {\color[HTML]{000000} 21.53±0.94}       & {\color[HTML]{000000} 17.86±0.22}       & {\color[HTML]{000000} 11.15±0.24}        & {\color[HTML]{000000} 17.33±0.54}       & {\color[HTML]{000000} 25.10±0.01}        & {\color[HTML]{000000} 22.04+2.23}       & {\color[HTML]{FF0000} 28.15±1.92}    \\
\cellcolor[HTML]{FFFFFF}{\color[HTML]{000000} }                                                        & {\color[HTML]{000000} ARI}                                                                            & {\color[HTML]{000000} 18.33±1.79}       & {\color[HTML]{000000} 20.50±0.51}        & {\color[HTML]{000000} 16.57±0.31}        & {\color[HTML]{000000} 21.63±1.49}      & {\color[HTML]{000000} 11.87±0.23}       & {\color[HTML]{000000} 20.39±0.70}         & {\color[HTML]{000000} 17.12±1.46}       & {\color[HTML]{000000} 13.13±0.71}       & {\color[HTML]{000000} 09.50±0.25}        & {\color[HTML]{000000} 13.62±0.57}       & {\color[HTML]{0000FF} 21.76±0.01}        & {\color[HTML]{000000} 14.74+1.99}       & {\color[HTML]{FF0000} 25.52±2.09}    \\
\multirow{-4}{*}{\cellcolor[HTML]{FFFFFF}{\color[HTML]{000000} \textbf{UAT}}}                          & {\color[HTML]{000000} F1}                                                                             & {\color[HTML]{000000} 47.95±1.52}       & {\color[HTML]{0000FF} 50.33±0.64}        & {\color[HTML]{000000} 50.26±0.16}        & {\color[HTML]{000000} 45.59±3.54}      & {\color[HTML]{000000} 25.79±0.29}       & {\color[HTML]{000000} 50.15±0.73}         & {\color[HTML]{000000} 39.44±2.19}       & {\color[HTML]{000000} 34.94±0.87}       & {\color[HTML]{000000} 35.18±0.32}        & {\color[HTML]{000000} 36.52±0.89}       & {\color[HTML]{000000} 45.69±0.08}        & {\color[HTML]{000000} 39.30+1.82}       & {\color[HTML]{FF0000} 55.24±1.69}    \\ \hline
\end{tabular}}
\caption{The average clustering performance of ten runs on six benchmark datasets. The performance is evaluated by four metrics with mean value and standard deviation. The {\color[HTML]{FF0000}red} and {\color[HTML]{0000FF}blue} values indicate the best and the runner-up results, respectively.}
\label{comparision_exper}
\end{table*}

\subsection{Objective Function}
The proposed method jointly optimizes two objectives including the positive sample loss $\mathcal{L}_{pos}$ and the negative sample loss $\mathcal{L}_{neg}$.

In detail, $\mathcal{L}_{pos}$ is the Mean Squared Error (MSE) loss between the normalized cross-view positive sample embeddings, as formulated: 

\begin{equation}
\begin{aligned}
\mathcal{L}_{pos} &= \frac{1}{K} \sum_{p=1}^K \sum_{i=1}^{n_p} \left \| \textbf{B}^{v_1}_{p[i,:]} - \textbf{B}^{v_2}_{p[i,:]} \right \|^2_2 \\
&= \frac{1}{K} \sum_{p=1}^K \sum_{i=1}^{n_p} \left (2 - 2 \left \langle \textbf{B}^{v_1}_{p[i,:]}, \textbf{B}^{v_2}_{p[i,:]} \right \rangle \right),
\end{aligned}
\label{pos_loss}
\end{equation}
where $\textbf{B}^{v_1}_{p[i,:]}$ and $\textbf{B}^{v_2}_{p[i,:]}$ denotes the $i$-th normalized node embedding in the $p$-th cluster of the first and second view, respectively. Besides, $n_p$ is the number of high-confidence samples in the $p$-th cluster. In this manner, the positive samples are pulled together. Besides, we define $\mathcal{L}_{neg}$ as the cosine similarity between different centers of the high-confidence embeddings:

\begin{equation}
\begin{aligned}
\mathcal{L}_{neg} &= \frac{1}{K^2-K} \sum_{p=1}^K \sum_{q=1}^K \frac{\left \langle \textbf{CEN}_p^{v_1}, \textbf{CEN}_q^{v_2} \right \rangle }{\| \textbf{CEN}_p^{v_1} \|_2 \cdot \| \textbf{CEN}_q^{v_2} \|_2 }, p \neq q,
\end{aligned}
\label{neg_loss}
\end{equation}
where $\textbf{CEN}_p^{v_1}$ is the $p$-th high-confidence center in the first view and $\textbf{CEN}_q^{v_2}$ is the $q$-th high-confidence center in the second view. By setting this, we push the negative samples away.

In summary, the total loss of the proposed CCGC is calculated as:

\begin{equation}
\begin{aligned}
\mathcal{L} = \mathcal{L}_{pos} + \alpha \mathcal{L}_{neg},
\end{aligned}
\label{total_loss}
\end{equation}
where $\alpha$ is a trade-off between $\mathcal{L}_{pos}$ and $\mathcal{L}_{neg}$. The detailed learning process of CCGC is shown in Algorithm \ref{ALGORITHM}.

\section{Experiments}

\subsection{Benchmark Datasets}
The experiments are conducted on six widely-used benchmark datasets, including CORA \cite{AGE}, CITESEER \cite{AGE}, BAT \cite{SCGC,RETHINK}, EAT \cite{SCGC}, UAT \cite{SCGC}, AMAP \cite{DCRN}. The summarized information is shown in Table \ref{DATASET_INFO}.

\begin{table}[h]
\centering
\small
\scalebox{1.}{
\begin{tabular}{@{}cccccc@{}}
\toprule
\textbf{Dataset} & \textbf{Type} & \textbf{Sample} & \textbf{Dimension} & \textbf{Edge}  & \textbf{Class} \\ \midrule
\textbf{CORA}  & Graph   & 2708    & 1433       & 5429   & 7       \\
\textbf{CITESEER} & Graph    & 3327    & 3703      & 4732   & 6       \\
\textbf{AMAP} & Graph  & 7650   & 745       & 119081  & 8       \\
\textbf{BAT}    & Graph  & 131    & 81      & 1038  & 4       \\
\textbf{EAT}    & Graph & 399    & 203       & 5994 & 4       \\
\textbf{UAT} & Graph  & 1190   & 239       & 13599  & 4       \\\bottomrule
\end{tabular}}
\caption{Statistics summary of six datasets.}
\label{DATASET_INFO} 
\end{table}

\subsection{Experiment Setup}
The experimental environment contains one desktop computer with the Intel Core i7-7820x CPU, one NVIDIA GeForce RTX 2080Ti GPU, 64GB RAM, and the PyTorch deep learning platform. The max training epoch number is set to 400. We minimize the total loss in Eq. \eqref{total_loss} with widely-used Adam optimizer \cite{ADAM} and then perform K-means over the learned embeddings. To obtain reliable clustering, we adopt a two-stage training strategy. The discriminative capacity of the model can be improved in the first stage. In the second stage, the contrastive learning mechanism can be enhanced by the high-confidence clustering pseudo labels. Ten runs are conducted for all methods. For the baselines, we adopt their source with original settings and reproduce the results. The hyper-parameter settings are summarized in Table 1 of the Appendix. The clustering performance is evaluated by four metrics including ACC, NMI, ARI, and F1 \cite{ZHOU_1,siwei_1,siwei_2,liang_1}.


\begin{figure*}[]
\footnotesize
\begin{minipage}{0.139\linewidth}
\centerline{\includegraphics[width=\textwidth]{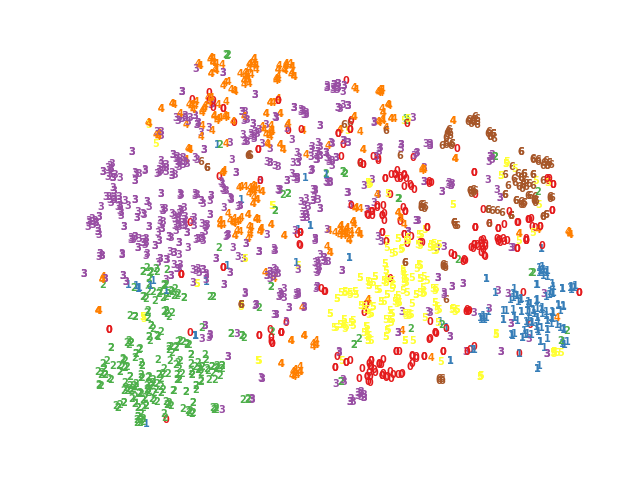}}
\vspace{3pt}
\centerline{\includegraphics[width=\textwidth]{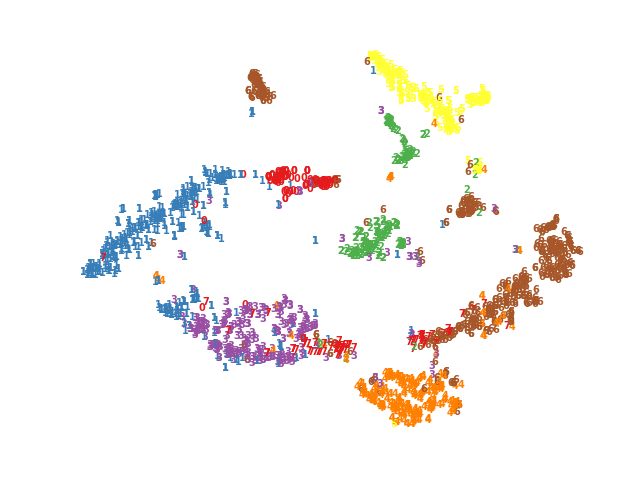}}
\vspace{3pt}
\centerline{DAEGC}
\end{minipage}
\begin{minipage}{0.139\linewidth}
\centerline{\includegraphics[width=\textwidth]{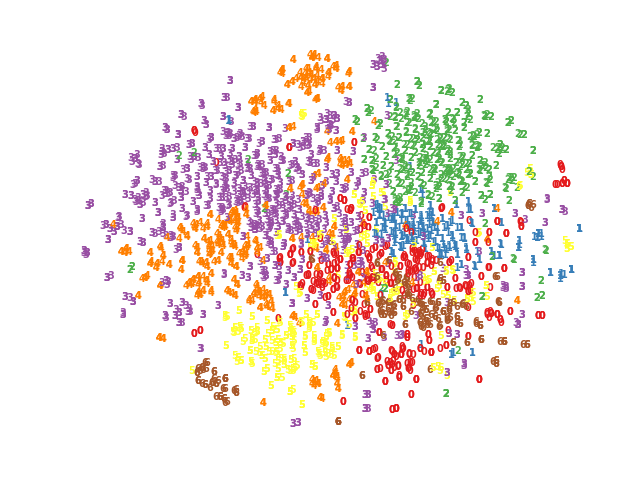}}
\vspace{3pt}
\centerline{\includegraphics[width=\textwidth]{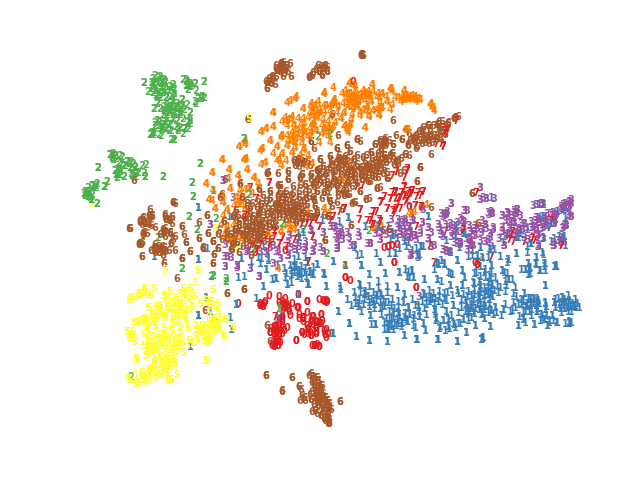}}
\vspace{3pt}
\centerline{MVGRL}
\end{minipage}
\begin{minipage}{0.139\linewidth}
\centerline{\includegraphics[width=\textwidth]{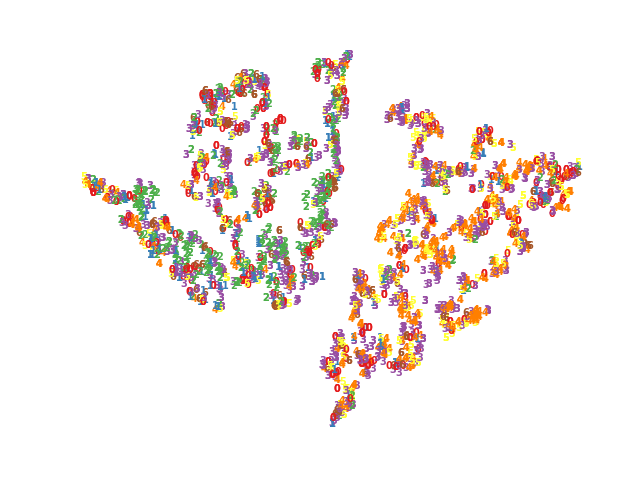}}
\vspace{3pt}
\centerline{\includegraphics[width=\textwidth]{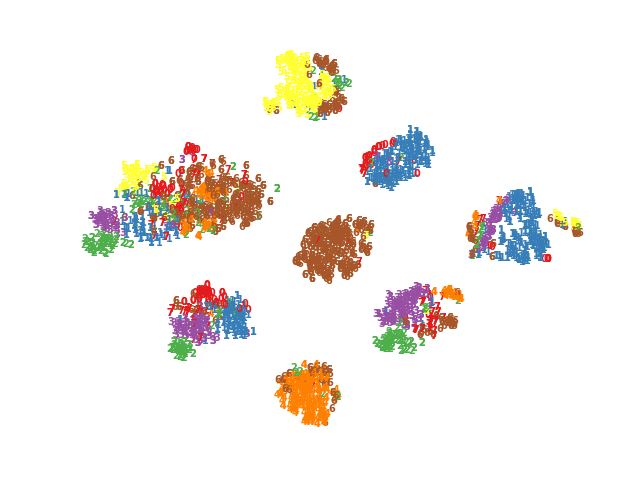}}
\vspace{3pt}
\centerline{SDCN}
\end{minipage}
\begin{minipage}{0.139\linewidth}
\centerline{\includegraphics[width=\textwidth]{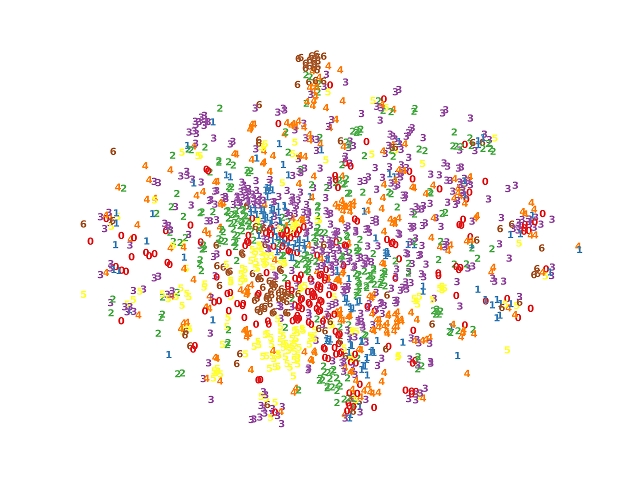}}
\vspace{3pt}
\centerline{\includegraphics[width=\textwidth]{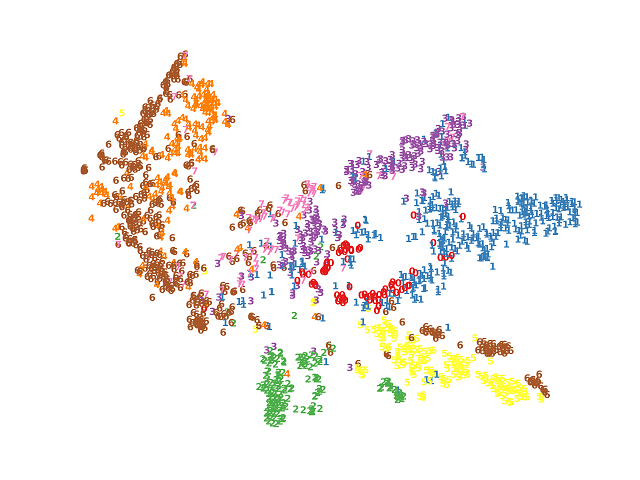}}
\vspace{3pt}
\centerline{AutoSSL}
\end{minipage}
\begin{minipage}{0.139\linewidth}
\centerline{\includegraphics[width=\textwidth]{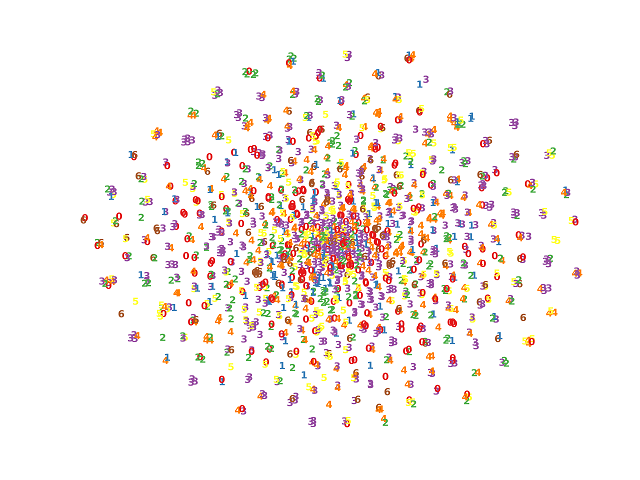}}
\vspace{3pt}
\centerline{\includegraphics[width=\textwidth]{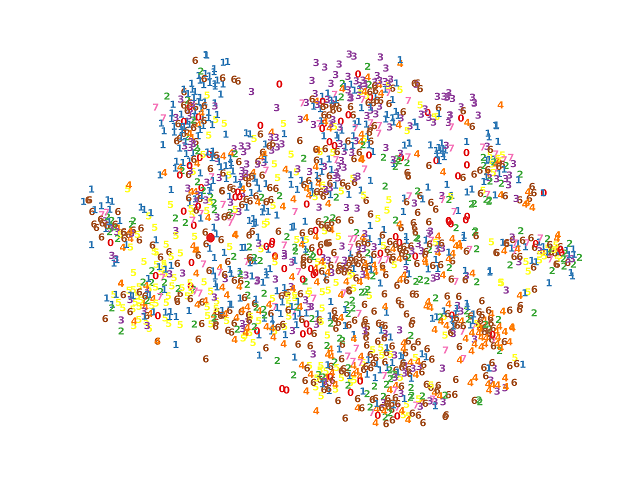}}
\vspace{3pt}
\centerline{AFGRL}
\end{minipage}
\begin{minipage}{0.139\linewidth}
\centerline{\includegraphics[width=\textwidth]{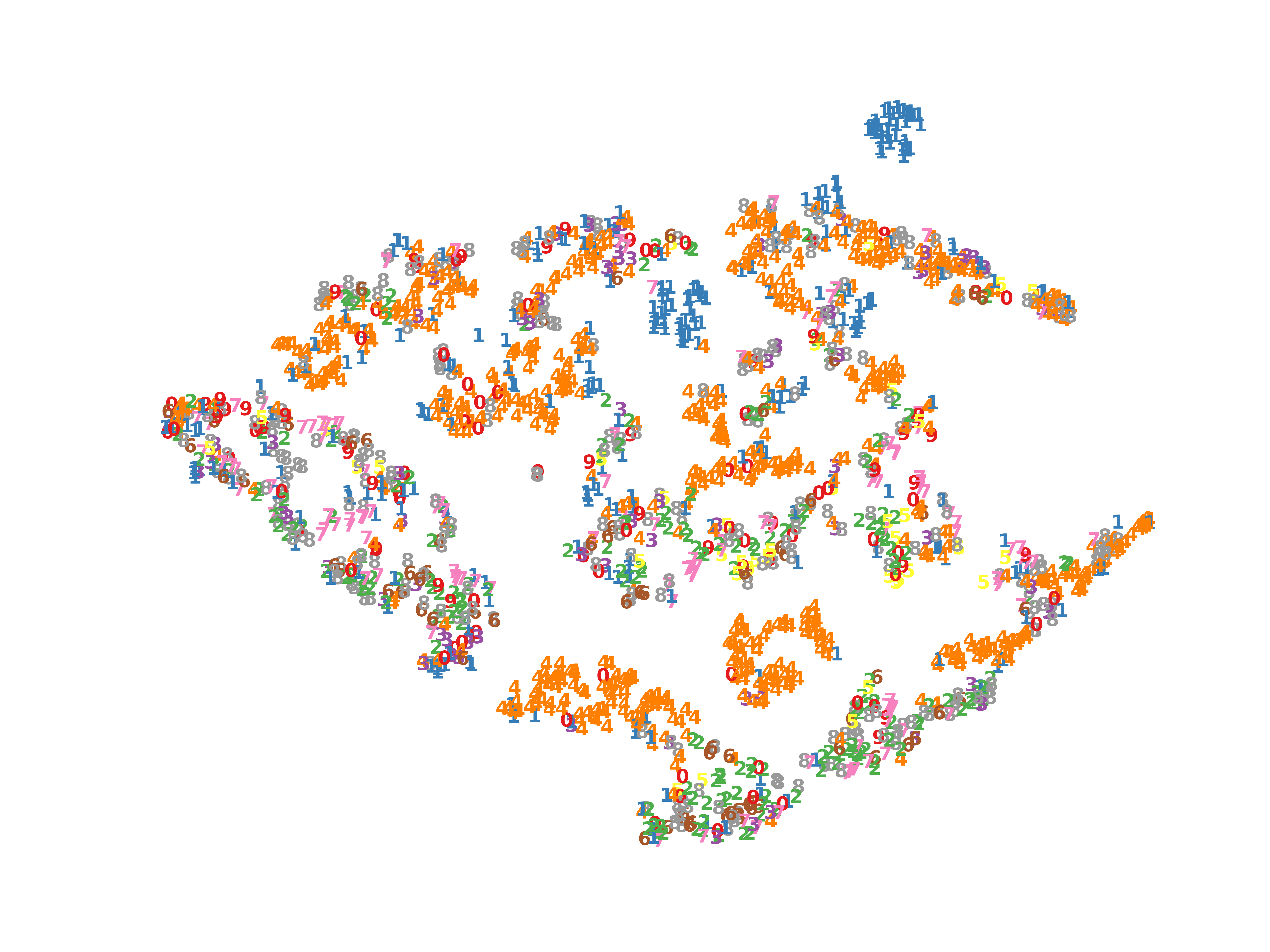}}
\vspace{3pt}
\centerline{\includegraphics[width=\textwidth]{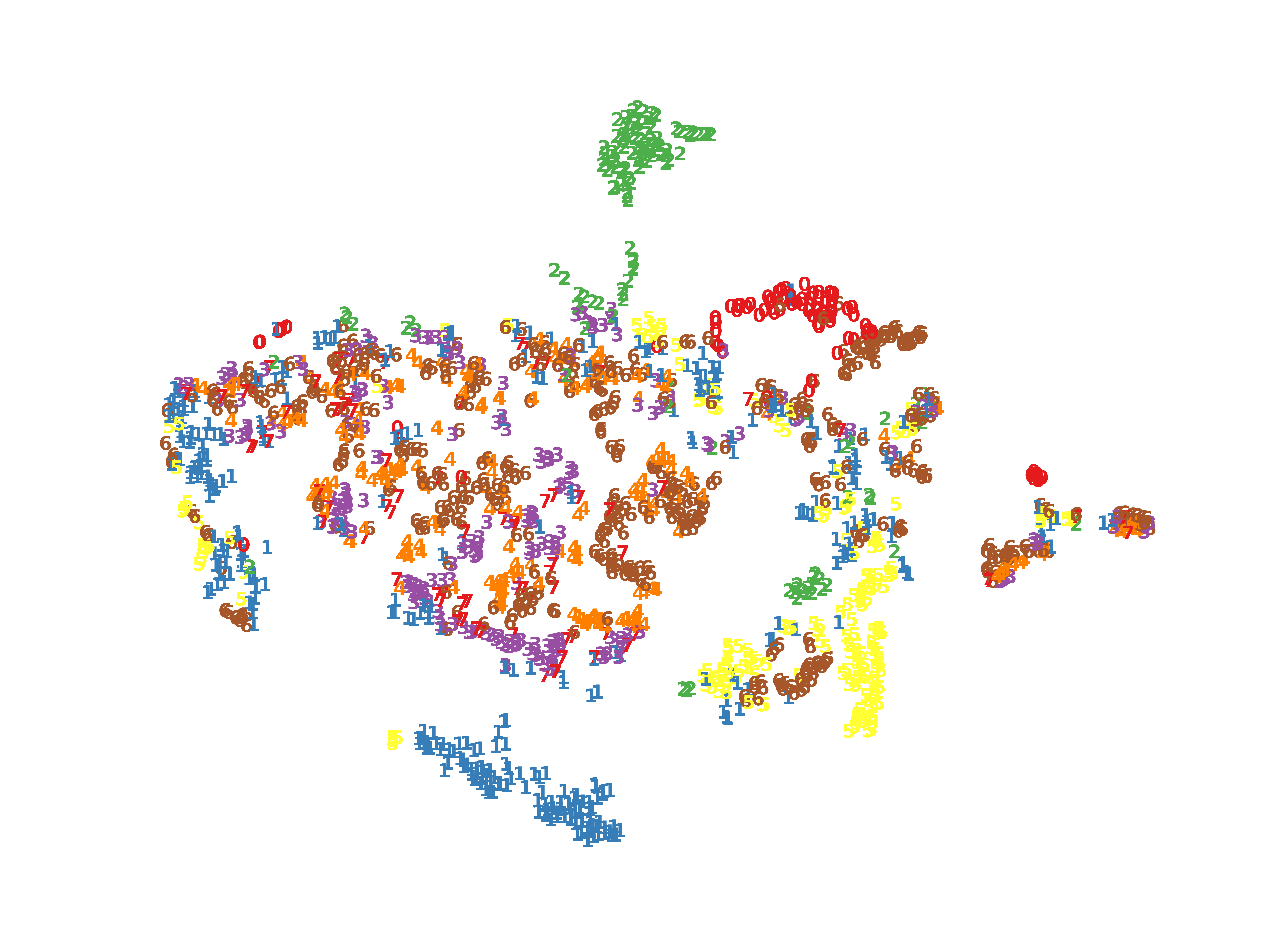}}
\vspace{3pt}
\centerline{GDCL}
\end{minipage}
\begin{minipage}{0.139\linewidth}
\centerline{\includegraphics[width=\textwidth]{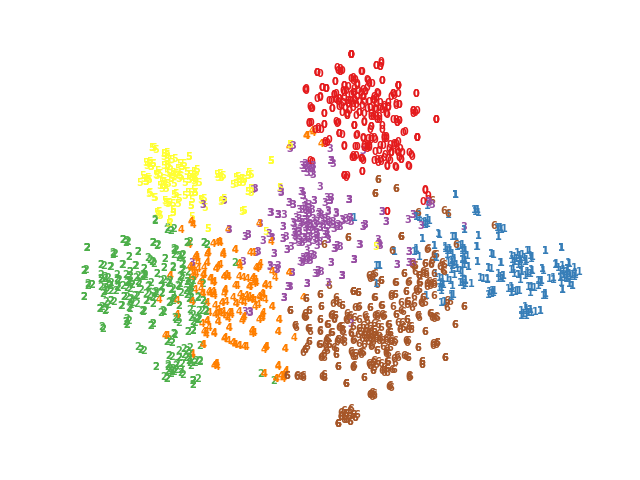}}
\vspace{3pt}
\centerline{\includegraphics[width=\textwidth]{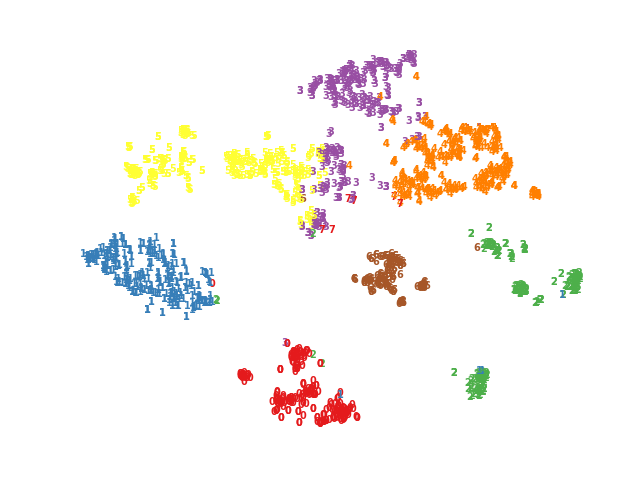}}
\vspace{3pt}
\centerline{Ours}
\end{minipage}

\caption{2D visualization on two datasets. The first row and second row correspond to CORA and AMAP, respectively.}
\label{t_SNE}  
\end{figure*}

\begin{algorithm}[]
\small
\caption{\textbf{CCGC}}
\label{ALGORITHM}
\flushleft{\textbf{Input}: The input graph $\mathcal{G}=\{\textbf{X},\textbf{A}\}$; The iteration number $I$; Hyper-parameters $\tau, t,\alpha$.} \\
\flushleft{\textbf{Output}: The clustering result \textbf{R}.} 
\begin{algorithmic}[1]
\STATE Obtain the smoothed attributes $\widetilde{\textbf{X}}$ with Eq \eqref{filter}.
\FOR{$i=1$ to $I$}
\STATE Encode $\widetilde{\textbf{X}}$ into two views with parameter un-shared Siamese encoders with Eq. \eqref{ENCODER}.
\STATE Normalize the embeddings $\textbf{E}^{v_1}, \textbf{E}^{v_2}$ with Eq. \eqref{normal}.
\STATE Perform K-means on \textbf{E} to obtain the clustering result.
\STATE Fuse $\textbf{E}^{v_1}$ and $\textbf{E}^{v_2}$ to obtain $\textbf{E}$ with Eq. \eqref{FUSION}.
\STATE Obtain high-confidence samples $\textbf{H}^{v_1}$ and $\textbf{H}^{v_2}$ with Eq. \eqref{two view}.
\STATE Construct positive and negative samples by DPS and RNS.
\STATE Calculate the positive samples loss $\mathcal{L}_{pos}$ with Eq. \eqref{pos_loss}.
\STATE Calculate the negative samples loss $\mathcal{L}_{neg}$ with Eq. \eqref{neg_loss}.
\STATE Update the whole network by minimizing $\mathcal{L}$ in Eq. \eqref{total_loss}.
\ENDFOR
\STATE Perform K-means on \textbf{E} to obtain the final clustering result \textbf{R}. 
\STATE \textbf{return} \textbf{R}
\end{algorithmic}
\end{algorithm}

\subsection{Performance Comparison}
In this subsection, we compare the clustering performance of our proposed algorithm with baselines on six datasets with four metrics. Among these methods, five classical deep graph clustering methods \cite{MGAE,DAEGC,ARGA,SDCN,DFCN} utilize the graph auto-encoder to learn the node representation for clustering. Besides, seven contrastive deep graph clustering methods \cite{AGE,MVGRL,AutoSSL,AGC-DRR, AFGRL,GDCL,ProGCL} improve the discriminative capability of samples by the contrastive strategies.

From the results in Table.\ref{comparision_exper}, we find that CCGC obtains better performance compared with the classical deep graph clustering methods. The reason is that contrastive learning could assist the model capture the supervision information implicitly. Besides, the contrastive methods achieve sub-optimal performance compared to ours. This is because we improve the discriminative capability and reliability of samples with the important clustering information. In summary, our method outperforms most of them on six datasets with four metrics. Taking the results on EAT dataset for example, CCGC exceeds the runner-up by 5.06\%, 9.92\%, 4.46\%, 4.34\% with respect to ACC, NMI, ARI, and F1. Besides, due to the limitation of the space, we conduct additional comparison experiments with nine baselines. Those results are shown in Table 2 of the Appendix. The results could also demonstrate the superiority of CCGC.


\subsection{Ablation Studies}
In this section, we first verify the effectiveness of two proposed sample construction strategies with experiments. Besides, we demonstrate the effect of parameter un-shared encoders and analyze the sensitivity of hyper-parameters in CCGC.

\begin{table*}[]
\centering
\scalebox{0.75}{
\begin{tabular}{c|c|ccccccc}
\hline
{\color[HTML]{000000} \textbf{Dataset}}                    & {\color[HTML]{000000} \textbf{Metric}} & {\color[HTML]{000000} \textbf{(w/o) Positive}} & {\color[HTML]{000000} \textbf{(w/o) Negitive}} & {\color[HTML]{000000} \textbf{Drop Edges}} & {\color[HTML]{000000} \textbf{Add Edges}} & {\color[HTML]{000000} \textbf{Diffusion}} & {\color[HTML]{000000} \textbf{Mask Feature}} & {\color[HTML]{000000} \textbf{Ours}}       \\ \hline
{\color[HTML]{000000} }                                    & {\color[HTML]{000000} ACC}             & {\color[HTML]{000000} 60.03±6.28}              & {\color[HTML]{000000} 70.29±0.86}              & {\color[HTML]{000000} 57.95±4.32}          & {\color[HTML]{000000} 57.89±3.16}         & {\color[HTML]{000000} 59.57±2.95}         & {\color[HTML]{000000} 67.40±1.76}            & {\color[HTML]{000000} \textbf{73.88±1.20}} \\
{\color[HTML]{000000} }                                    & {\color[HTML]{000000} NMI}             & {\color[HTML]{000000} 47.33±3.97}              & {\color[HTML]{000000} 53.67±1.15}              & {\color[HTML]{000000} 39.32±4.90}          & {\color[HTML]{000000} 39.11±3.82}         & {\color[HTML]{000000} 39.84±2.72}         & {\color[HTML]{000000} 48.84±2.02}            & {\color[HTML]{000000} \textbf{55.56±1.04}} \\
{\color[HTML]{000000} }                                    & {\color[HTML]{000000} ARI}             & {\color[HTML]{000000} 37.66±5.86}              & {\color[HTML]{000000} 47.09±1.29}              & {\color[HTML]{000000} 29.10±4.52}          & {\color[HTML]{000000} 29.74±2.84}         & {\color[HTML]{000000} 30.73±2.95}         & {\color[HTML]{000000} 41.32±2.33}            & {\color[HTML]{000000} \textbf{52.51±1.89}} \\
\multirow{-4}{*}{{\color[HTML]{000000} \textbf{CORA}}}     & {\color[HTML]{000000} F1}              & {\color[HTML]{000000} 53.85±9.38}              & {\color[HTML]{000000} 68.48±1.17}              & {\color[HTML]{000000} 53.45±5.57}          & {\color[HTML]{000000} 55.33±   6.39}      & {\color[HTML]{000000} 55.01±7.22}         & {\color[HTML]{000000} 63.16±3.54}            & {\color[HTML]{000000} \textbf{70.98±2.79}} \\ \hline
{\color[HTML]{000000} }                                    & {\color[HTML]{000000} ACC}             & {\color[HTML]{000000} 58.26±6.19}              & {\color[HTML]{000000} 67.92±1.32}              & {\color[HTML]{000000} 66.55±1.27}          & {\color[HTML]{000000} 66.31±1.40}         & {\color[HTML]{000000} 68.32±0.62}         & {\color[HTML]{000000} 69.14±0.66}            & {\color[HTML]{000000} \textbf{69.84±0.94}} \\
{\color[HTML]{000000} }                                    & {\color[HTML]{000000} NMI}             & {\color[HTML]{000000} 38.62±3.34}              & {\color[HTML]{000000} 42.05±1.34}              & {\color[HTML]{000000} 38.91±2.04}          & {\color[HTML]{000000} 39.43±1.72}         & {\color[HTML]{000000} 41.83±0.94}         & {\color[HTML]{000000} 42.49±0.89}            & {\color[HTML]{000000} \textbf{44.33±0.79}} \\
{\color[HTML]{000000} }                                    & {\color[HTML]{000000} ARI}             & {\color[HTML]{000000} 35.78±4.28}              & {\color[HTML]{000000} 42.66±1.65}              & {\color[HTML]{000000} 38.85±1.63}          & {\color[HTML]{000000} 39.00±1.32}         & {\color[HTML]{000000} 41.23±0.99}         & {\color[HTML]{000000} 43.12±1.16}            & {\color[HTML]{000000} \textbf{44.68±1.80}} \\
\multirow{-4}{*}{{\color[HTML]{000000} \textbf{CITESEER}}} & {\color[HTML]{000000} F1}              & {\color[HTML]{000000} 45.90±7.76}              & {\color[HTML]{000000} \textbf{62.82±0.92}}     & {\color[HTML]{000000} 58.38±1.51}          & {\color[HTML]{000000} 59.56±1.40}         & {\color[HTML]{000000} 59.89±0.83}         & {\color[HTML]{000000} 60.78±1.43}            & {\color[HTML]{000000} 62.71±2.06}          \\ \hline
{\color[HTML]{000000} }                                    & {\color[HTML]{000000} ACC}             & {\color[HTML]{000000} 29.81±1.71}              & {\color[HTML]{000000} 73.74±0.82}              & {\color[HTML]{000000} 75.84±1.12}          & {\color[HTML]{000000} 75.75±1.77}         & {\color[HTML]{000000} 70.89±3.28}         & {\color[HTML]{000000} 76.48±1.88}            & {\color[HTML]{000000} \textbf{77.25±0.41}} \\
{\color[HTML]{000000} }                                    & {\color[HTML]{000000} NMI}             & {\color[HTML]{000000} 15.18±1.93}              & {\color[HTML]{000000} 62.65±1.48}              & {\color[HTML]{000000} 62.98±1.15}          & {\color[HTML]{000000} 63.11±1.89}         & {\color[HTML]{000000} 58.14±2.05}         & {\color[HTML]{000000} 66.44±0.85}            & {\color[HTML]{000000} \textbf{67.44±0.48}} \\
{\color[HTML]{000000} }                                    & {\color[HTML]{000000} ARI}             & {\color[HTML]{000000} 05.85±0.95}              & {\color[HTML]{000000} 52.74±1.51}              & {\color[HTML]{000000} 55.81±1.63}          & {\color[HTML]{000000} 55.79±2.85}         & {\color[HTML]{000000} 49.68±3.16}         & {\color[HTML]{000000} 56.81±2.09}            & {\color[HTML]{000000} \textbf{57.99±0.66}} \\
\multirow{-4}{*}{{\color[HTML]{000000} \textbf{AMAP}}}     & {\color[HTML]{000000} F1}              & {\color[HTML]{000000} 26.61±3.49}              & {\color[HTML]{000000} 68.45±1.08}              & {\color[HTML]{000000} 69.82±3.24}          & {\color[HTML]{000000} 69.83±3.13}         & {\color[HTML]{000000} 60.43±6.16}         & {\color[HTML]{000000} 71.13±1.40}            & {\color[HTML]{000000} \textbf{72.18±0.57}} \\ \hline
{\color[HTML]{000000} }                                    & {\color[HTML]{000000} ACC}             & {\color[HTML]{000000} 65.19±2.00}              & {\color[HTML]{000000} 73.59±2.32}              & {\color[HTML]{000000} 50.00±3.87}          & {\color[HTML]{000000} 67.02±2.71}         & {\color[HTML]{000000} 52.60±2.72}         & {\color[HTML]{000000} 72.06±2.92}            & {\color[HTML]{000000} \textbf{75.04±1.78}} \\
{\color[HTML]{000000} }                                    & {\color[HTML]{000000} NMI}             & {\color[HTML]{000000} 44.08±1.35}              & {\color[HTML]{000000} 47.07±2.38}              & {\color[HTML]{000000} 24.59±2.01}          & {\color[HTML]{000000} 45.13±4.73}         & {\color[HTML]{000000} 29.78±5.06}         & {\color[HTML]{000000} 48.67±2.47}            & {\color[HTML]{000000} \textbf{50.23±2.43}} \\
{\color[HTML]{000000} }                                    & {\color[HTML]{000000} ARI}             & {\color[HTML]{000000} 38.46±2.60}              & {\color[HTML]{000000} 44.37±3.47}              & {\color[HTML]{000000} 18.91±3.80}          & {\color[HTML]{000000} 39.18±5.16}         & {\color[HTML]{000000} 21.12±5.70}         & {\color[HTML]{000000} 44.07±3.80}            & {\color[HTML]{000000} \textbf{46.95±3.09}} \\
\multirow{-4}{*}{{\color[HTML]{000000} \textbf{BAT}}}      & {\color[HTML]{000000} F1}              & {\color[HTML]{000000} 62.24±2.57}              & {\color[HTML]{000000} 73.36±2.35}              & {\color[HTML]{000000} 48.44±4.90}          & {\color[HTML]{000000} 65.54±2.66}         & {\color[HTML]{000000} 48.72±2.54}         & {\color[HTML]{000000} 71.57±3.46}            & {\color[HTML]{000000} \textbf{74.90±1.80}} \\ \hline
{\color[HTML]{000000} }                                    & {\color[HTML]{000000} ACC}             & {\color[HTML]{000000} 48.42±2.91}              & {\color[HTML]{000000} 52.31±1.66}              & {\color[HTML]{000000} 45.76±1.00}          & {\color[HTML]{000000} 49.32±1.53}         & {\color[HTML]{000000} 45.71±1.67}         & {\color[HTML]{000000} 52.61±1.70}            & {\color[HTML]{000000} \textbf{57.19±0.66}} \\
{\color[HTML]{000000} }                                    & {\color[HTML]{000000} NMI}             & {\color[HTML]{000000} 25.88±2.73}              & {\color[HTML]{000000} 26.12±2.07}              & {\color[HTML]{000000} 14.57±3.14}          & {\color[HTML]{000000} 23.45±2.00}         & {\color[HTML]{000000} 19.98±1.84}         & {\color[HTML]{000000} 27.18±1.55}            & {\color[HTML]{000000} \textbf{33.85±0.87}} \\
{\color[HTML]{000000} }                                    & {\color[HTML]{000000} ARI}             & {\color[HTML]{000000} 19.29±2.02}              & {\color[HTML]{000000} 20.94±1.40}              & {\color[HTML]{000000} 10.43±1.63}          & {\color[HTML]{000000} 16.85±1.76}         & {\color[HTML]{000000} 14.71±3.32}         & {\color[HTML]{000000} 20.50±1.46}            & {\color[HTML]{000000} \textbf{27.71±0.41}} \\
\multirow{-4}{*}{{\color[HTML]{000000} \textbf{EAT}}}      & {\color[HTML]{000000} F1}              & {\color[HTML]{000000} 46.39±4.98}              & {\color[HTML]{000000} 52.63±1.90}              & {\color[HTML]{000000} 43.78±1.36}          & {\color[HTML]{000000} 48.99±2.18}         & {\color[HTML]{000000} 40.46±3.81}         & {\color[HTML]{000000} 52.51±2.02}            & {\color[HTML]{000000} \textbf{57.09±0.94}} \\ \hline
{\color[HTML]{000000} }                                    & {\color[HTML]{000000} ACC}             & {\color[HTML]{000000} 41.39±2.96}              & {\color[HTML]{000000} 49.04±1.33}              & {\color[HTML]{000000} 57.02±1.44}          & {\color[HTML]{000000} 55.61±1.00}         & {\color[HTML]{000000} 51.45±2.04}         & {\color[HTML]{000000} 52.33±1.95}            & {\color[HTML]{000000} \textbf{56.34±1.11}} \\
{\color[HTML]{000000} }                                    & {\color[HTML]{000000} NMI}             & {\color[HTML]{000000} 12.08±1.86}              & {\color[HTML]{000000} 22.89±1.71}              & {\color[HTML]{000000} 26.07±1.43}          & {\color[HTML]{000000} 29.06±1.49}         & {\color[HTML]{000000} 24.18±2.19}         & {\color[HTML]{000000} 24.12±1.99}            & {\color[HTML]{000000} \textbf{28.15±1.92}} \\
{\color[HTML]{000000} }                                    & {\color[HTML]{000000} ARI}             & {\color[HTML]{000000} 7.70±0.63}               & {\color[HTML]{000000} 20.70±1.22}              & {\color[HTML]{000000} \textbf{26.03±1.88}} & {\color[HTML]{000000} 23.69±1.58}         & {\color[HTML]{000000} 22.55±2.80}         & {\color[HTML]{000000} 18.54±3.07}            & {\color[HTML]{000000} 25.52±2.09}          \\
\multirow{-4}{*}{{\color[HTML]{000000} \textbf{UAT}}}      & {\color[HTML]{000000} F1}              & {\color[HTML]{000000} 36.14±3.32}              & {\color[HTML]{000000} 44.22±2.24}              & {\color[HTML]{000000} 54.67±1.67}          & {\color[HTML]{000000} 54.74± 1.16}        & {\color[HTML]{000000} 46.06±2.97}         & {\color[HTML]{000000} 51.17±2.16}            & {\color[HTML]{000000} \textbf{55.24±1.69}} \\ \hline
\end{tabular}}
\caption{Ablation studies of CCGC on six datasets.}
\label{ablation_study}
\end{table*}

\subsubsection{\textbf{Effectiveness of DPS and RNS}} \label{pos_neg_ablation}
To verify the effect of the proposed Discriminative Positive sample construction Strategy (DPS) and Reliable Negative sample construction Strategy (RNS), we conduct extensive experiments as shown in Table \ref{ablation_study}. For simplicity, we denote ``(w/o) DPS'' and ``(w/o) RNS'' as replacing DPS and RNS in our model with the regular positive and negative sample construction fashion \cite{DCRN}, i.e., regarding the same samples in two view as positive samples while considering other samples as negative ones. From the observations, we conclude that the performance will decrease without any one of DPS and RNS, revealing that both strategies make essential contributions to boosting the performance. In addition, the quality of positive and negative sample pairs is improved compared with the regular sample construction fashion. Overall, the experimental results have verified the effectiveness of DPS and RNS.

\subsubsection{\textbf{Effectiveness of Parameter Un-shared Encoders}} \label{augmentation}
To avoid the complex augmentations on graphs, we design the un-shared Simases encoders to conduct two-node views. In this part, we compare our view construction method with other classical graph data augmentations including edge dropping \cite{SCAGC}, edge adding \cite{SCAGC}, graph diffusion \cite{MVGRL}, and feature masking \cite{GDCL}. Concretely, in Table \ref{ablation_study}, we first make the encoders in CCGC to share parameter and then adopt the data augmentation as randomly dropping 20\% edges (``Drop Edges''), or randomly adding 20\% edges (``Add Edges''), or graph diffusion (``Diffusion'') with 0.20 teleportation rate, or randomly masking 20\% features (``Mask Features''). From the results, we observe that the commonly used graph augmentations might lead to semantic drift \cite{AFGRL}, thus undermining the performance. In summary, expensive experiments have demonstrated the effectiveness of the proposed parameter un-shared encoders.

\begin{figure}[h]
\centering
\small
\begin{minipage}{0.32\linewidth}
\centerline{\includegraphics[width=1\textwidth]{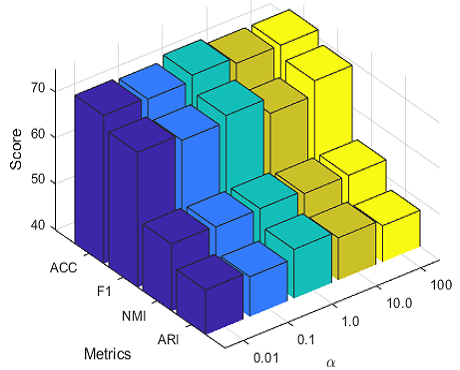}}
\vspace{3pt}
\textbf{\centerline{CORA}}
\centerline{\includegraphics[width=\textwidth]{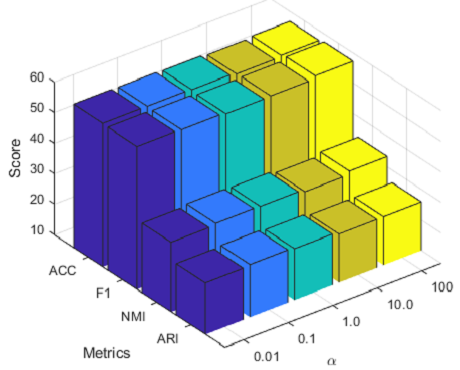}}
\vspace{3pt}
\textbf{\centerline{EAT}}
\end{minipage}
\begin{minipage}{0.32\linewidth}
\centerline{\includegraphics[width=\textwidth]{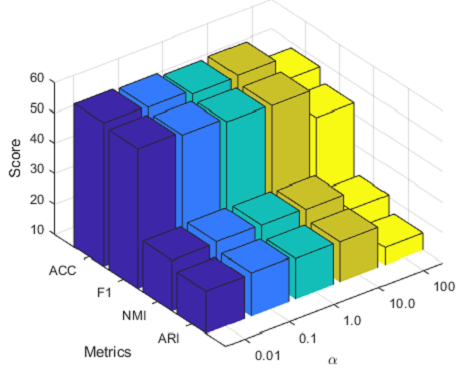}}
\vspace{3pt}
\textbf{\centerline{UAT}}
\centerline{\includegraphics[width=1\textwidth]{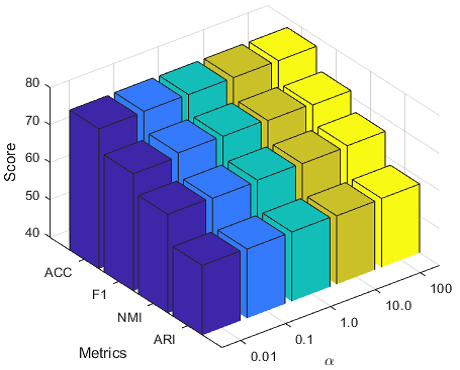}}
\vspace{3pt}
\textbf{\centerline{AMAP}}
\end{minipage}
\begin{minipage}{0.32\linewidth}
\centerline{\includegraphics[width=\textwidth]{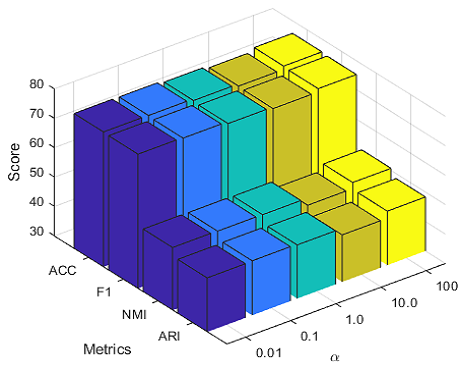}}
\vspace{3pt}
\textbf{\centerline{BAT}}
\centerline{\includegraphics[width=\textwidth]{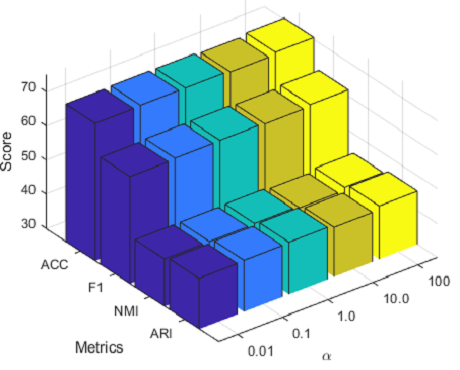}}
\vspace{3pt}
\textbf{\centerline{CITESEER}}
\end{minipage}
\caption{Sensitivity analysis of the hyper-parameter $\alpha$ on six datasets.}
\label{Sensitivity_alpha}
\end{figure}

\begin{figure}[h]
\centering
\small
\begin{minipage}{0.32\linewidth}
\centerline{\includegraphics[width=1\textwidth]{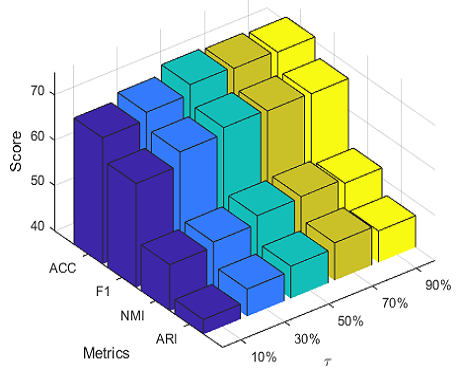}}
\vspace{3pt}
\textbf{\centerline{CORA}}
\centerline{\includegraphics[width=\textwidth]{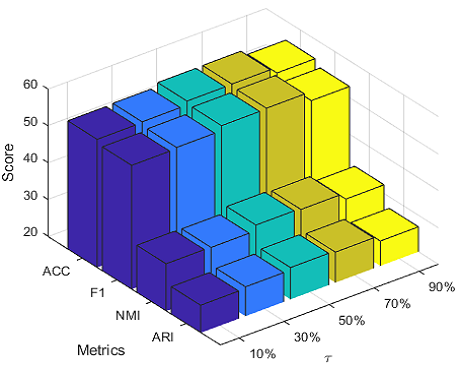}}
\vspace{3pt}
\textbf{\centerline{EAT}}
\end{minipage}
\begin{minipage}{0.32\linewidth}
\centerline{\includegraphics[width=\textwidth]{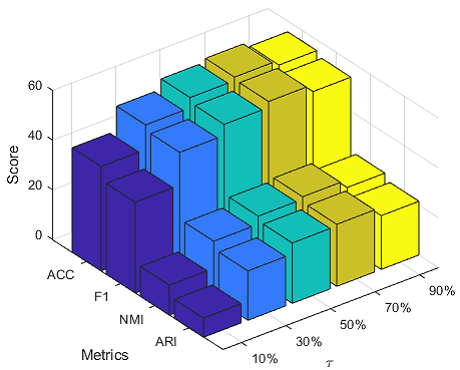}}
\vspace{3pt}
\textbf{\centerline{UAT}}
\centerline{\includegraphics[width=1\textwidth]{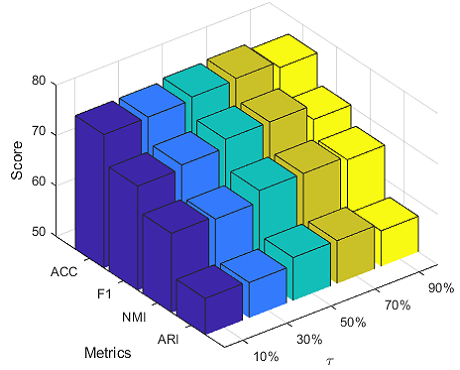}}
\vspace{3pt}
\textbf{\centerline{AMAP}}
\end{minipage}
\begin{minipage}{0.32\linewidth}
\centerline{\includegraphics[width=\textwidth]{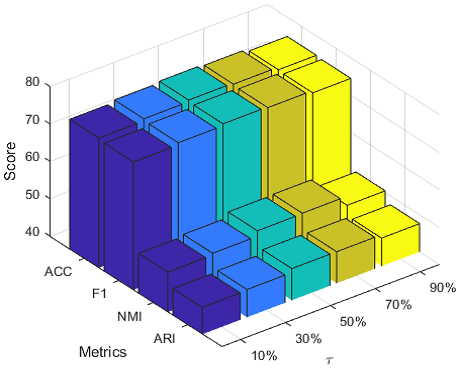}}
\vspace{3pt}
\textbf{\centerline{BAT}}
\vspace{3pt}
\centerline{\includegraphics[width=\textwidth]{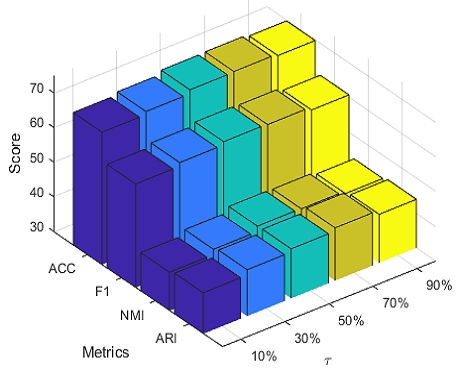}}
\vspace{3pt}
\textbf{\centerline{CITESEER}}
\end{minipage}
\caption{Sensitivity analysis of the hyper-parameter $\tau$ on six datasets.}
\label{Sensitivity_t}
\end{figure}

\subsection{Hyper-parameter Analysis}

\subsubsection{\textbf{Sensitivity Analysis of hyper-parameter threshold $\tau$}}
We investigate the influence of the hyper-parameter threshold $\tau$ on six datasets as shown in Fig.\ref{Sensitivity_t}. From the results, we observe that the model obtains promising performance when $\tau \in [50\%, 70\%]$. The reasons are as follows: 1) When $\tau < 50\%$, the discriminative capacity of the network is limited due to few number of positive samples. 2) When $\tau > 70\%$, the over-confidence pseudo labels would easily lead the network to confirmation bias \cite{confirmation}.

\subsubsection{\textbf{Sensitivity Analysis of hyper-parameter $\alpha$}}
Besides, to the trade-off hyper-parameter $\alpha$, the experimental results are shown in Fig.\ref{Sensitivity_alpha}. From these results, we observe that the performance will not fluctuate greatly when $\alpha$ is varying. This demonstrates that our CCGC is insensitive to $\alpha$. Moreover, CCGC is also insensitive to the layer number $t$ of Laplacian filters. Experimental evidences can be found in Fig. 1 in Appendix.

\subsection{Visualization Analysis} In this part, we visualize the distribution of the learned embeddings of six baselines and CCGC to show the superiority of CCGC on CORA and AMAP datasets via $t$-SNE algorithm \cite{T_SNE}. As shown in Fig. \ref{t_SNE}, we can conclude that CCGC better reveals the intrinsic clustering structure compared with other baselines.

\section{Conclusion}

In this work, we propose a Cluster-guided Contrastive deep Graph Clustering network termed CCGC to improve the quality of positive and negative samples. To be specific, we firstly construct two views with the un-shared parameters Siamese encoders to avoid semantic drift caused by the inappropriate graph data augmentations. Besides, the proposed positive and negative samples construction strategies improve the discriminative capability and reliability of samples by mining the supervision information in the high-confidence clustering pseudo labels. Extensive experiments on six datasets demonstrate the effectiveness of our proposed method.

\newpage

\section{Acknowledgments}
This work was supported by the National Key R\&D Program of China (project no. 2020AAA0107100, 2021YFB3100700) and the National Natural Science Foundation of China (project no. 61922088, 61976196, 62006237, and 61872371).


\bibliography{aaai23}

\end{document}